\documentclass[lettersize,journal]{IEEEtran}
\usepackage{amsmath,amsfonts}
\usepackage{algorithmic}
\usepackage{algorithm}
\usepackage{array}
\usepackage[caption=false,font=normalsize,labelfont=sf,textfont=sf]{subfig}
\usepackage{textcomp}
\usepackage{stfloats}
\usepackage{url}
\usepackage{verbatim}
\usepackage{graphicx}
\usepackage{cite}
\usepackage{anyfontsize}
\usepackage{hyperref}
\usepackage[table,xcdraw,dvipsnames]{xcolor}
\usepackage{breqn}
\usepackage{dsfont}
\usepackage{multicol}
\begin{document}

\noindent
© 2022 IEEE. Personal use of this material is permitted. Permission from IEEE must be obtained for all other uses, in any current or future media, including reprinting/republishing this material for advertising or promotional purposes, creating new collective works, for resale or redistribution to servers or lists, or reuse of any copyrighted component of this work in other works.
\\

\noindent
Journal: IEEE Journal of Selected Topics in Signal Processing\\
Publication Date: OCTOBER 2022\\
Volume: 16, Issue: 6\\
On Page(s): 1211-1226\\
Print ISSN: 1932-4553\\
Online ISSN: 1941-0484\\
Digital Object Identifier: 10.1109/JSTSP.2022.3206084

\newpage

\title{Self-supervised language learning from raw audio:\\ 
Lessons from the Zero Resource Speech Challenge}

\author{
Ewan Dunbar, Nicolas Hamilakis, 
and Emmanuel Dupoux
\thanks{Ewan Dunbar is with the Department of French,
University of Toronto, Toronto, ON, M5S 1A1, Canada (e-mail:
ewan.dunbar@utoronto.ca).

Nicolas Hamilakis is with the École Normale Supérieure, 75005 Paris, France, and with Inria, 75013 Paris, France (e-mail: nick.hamilakis562@gmail.com).

Emmanuel Dupoux is with the École des Hautes Études en Sciences Sociales, 75006 Paris, France, and with Meta AI, 75009 Paris, France (e-mail: emmanuel.dupoux@gmail.com).}}


\markboth{Journal of Selected Topics in Signal Processing}{Lessons from the Zero Resource Speech Challenge series}

\IEEEpubid{1941-0484~\copyright~2022 IEEE}

\maketitle

\IEEEpubidadjcol

\begin{abstract}
Recent progress in self-supervised or unsupervised machine learning has opened the possibility of building a full speech processing system from raw audio without using any textual representations or expert labels such as phonemes, dictionaries or parse trees. The contribution of the Zero Resource Speech Challenge series since 2015 has been to break down this long-term objective into four well-defined tasks---Acoustic Unit Discovery, Spoken Term Discovery, Discrete Resynthesis, and Spoken Language Modeling---and introduce associated metrics and benchmarks enabling model comparison and cumulative progress. We present an overview of the six editions of this challenge series since 2015, discuss the lessons learned, and outline the areas which need more work or give puzzling results. 
\end{abstract}


\section{Introduction}


\begin{figure*}[b]
\centering
\includegraphics[width=0.7\textwidth]{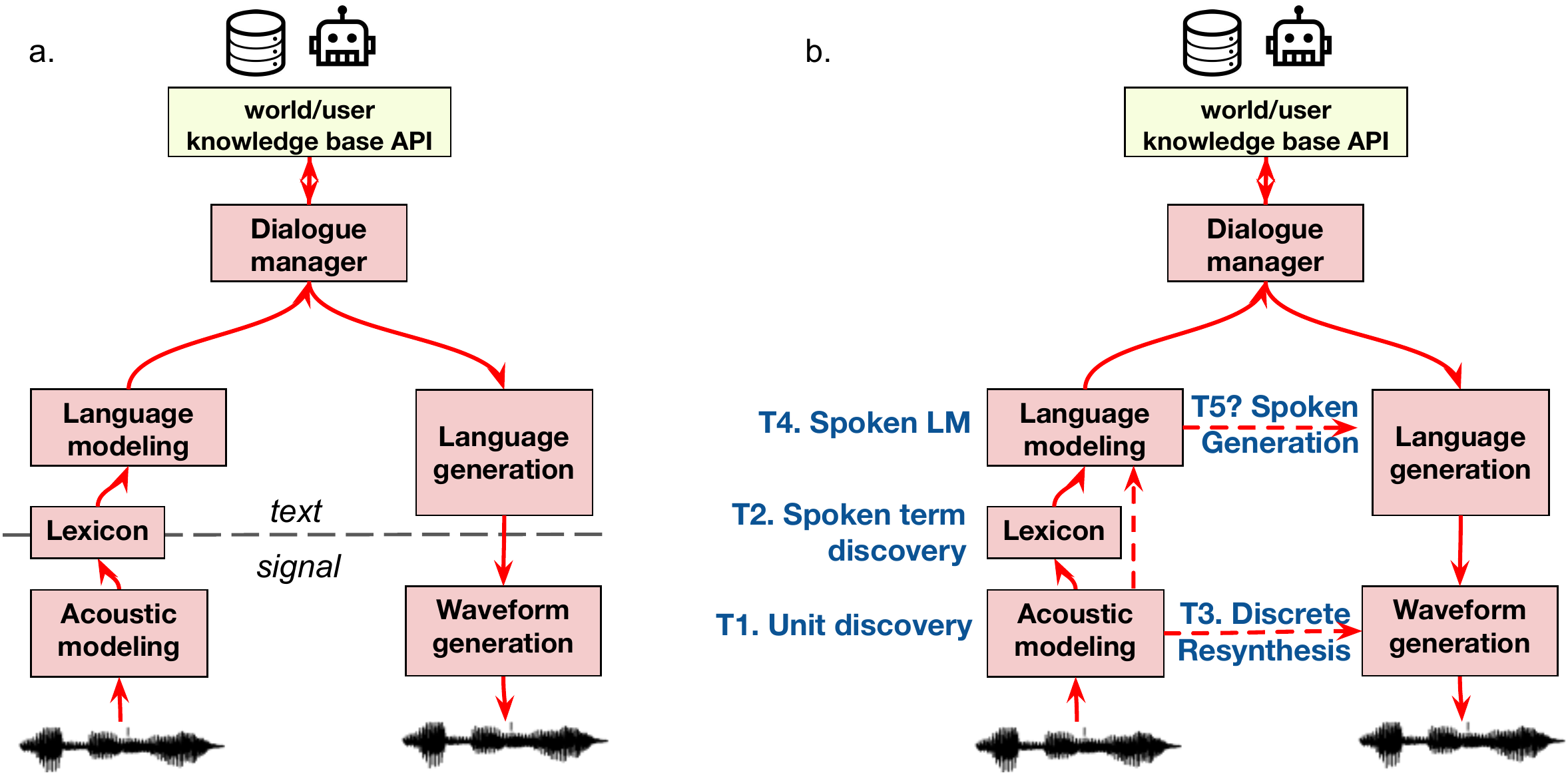}
\caption{a. Traditional pipeline for a spoken assistant based on textual resources b. Pipeline and tasks for the zero resource challenge series.}
\label{fig:blueprint}
\end{figure*}

\IEEEPARstart{F}{or} a long time, language technology has been developed principally using large quantities of textual resources. This makes sense, since, as far as technological applications are concerned, language has primarily been used in written form. When it comes to dealing with spoken  language, however, this has given rise to a division of labor between, on the one hand, speech components which aim at converting speech to text or text to speech (ASR, automatic speech recognition, and TTS, text-to-speech synthesis), and, on the other hand,  components that perform a variety of language tasks based on text (language understanding, dialogue, language generation). As a result, even speech-first applications like speech-to-speech translation or speech assistants like Alexa or Siri are cobbled together in a Frankensteinian fashion, with some components trained on text and others trained on speech (see Figure \ref{fig:blueprint}a)---and with all the speech components trained using large amounts of supervision (textual transcription) so that they can communicate with the text-based components. But is this a necessity? Could we build spoken-language based applications directly from the audio stream without using any text?

\begin{table*}[]
\centering

\caption{\textbf{Summary of the metrics used in the Zero Resource Challenge series for each task}.}

\def\arraystretch{1.2} 
\begin{tabular}{p{3.2cm}p{1.5cm}p{2.99cm}p{4.1cm}p{4.2cm}}
\hline
\bf Task                   & \bf Ling. Level               & \bf Model outputs examined      & \bf Metric     &\bf Example   \\
\hline
T1. Unit Discovery         & Phonetic                      & sequences of frame\newline embeddings   & \textbf{ABX}: $d(a,x)<d(b,x)?\newline a\in A, b\in B, x\neq a\in A $                         & $a=$'fly'${_1}$, $b=$'fry'${_1}$, $x=$'fly'${_2}$ \\ \hline
                           & Lexical (matching)            & pairs of segments      & \textbf{NED} : $ED(a,b)/max(|a|,|b|)$ \newline \textbf{COV}: fraction of corpus covered      & $a=$'rose', $b=$'prose'\\  \cline{2-5}
T2. Spoken Term Discovery  & Lexical (clustering)          & cluster memberships of\newline segments & \textbf{Grouping F-score} \newline \textbf{Type F-score}           & c$_1$=\{'rose'$_1$, 'prose'$_1$, 'ose'$_2$\}\newline c$_2$=\{'rose'$_2$,'rose'$_3$\} \\ \cline{2-5} 
                           & Lexical\newline(segmentation) &  list of time-stamps            & \textbf{Token F-score}: as defined in text  \newline \textbf{Boundary F-score}: segmentation & 'Ho$|$ware$|$youto$|$day' \\ \hline 
T3. Unsup. Discrete Resynthesis ("TTS without T") & Phonetic                      & series of discrete units\newline waveforms& \textbf{Bitrate}: $\frac{n}{D(U)}\sum{p(u_i)log( p(u_i))}$ \newline \textbf{MOS}: human evaluation        & $U=u_{1},..,u_{n}$\newline $U=$'How are you today'\\ \hline
                           & Lexical\newline(frequency)    & (pseudo) probabilities          & \textbf{spot-the-word}:  $\hat{p}(a)\textgreater{}\hat{p}(b)?$     & $a=$'brick', b=$^*$'blick'\\  \cline{2-5}
T4. Spoken LM              & Lexical\newline(semantics)    & word embeddings                 & \textbf{similarity}:  $d(a,b) \propto d_{h}(a,b) ?$                & $a=$'abduct', b='kidnap'\\  \cline{2-5}
                           & Syntax                        & (pseudo) probabilities          & \textbf{accept. judgment}:  $\hat{p}(a)\textgreater{}\hat{p}(b)?$  & $a$='dogs sleep', $b$=$^*$'dogs sleeps'\\ \hline
\multicolumn{5}{p{17cm}}{\,\newline \textit{Note.} $d$ is a dissimilarity measure between word or frame embeddings, $d_h$ is a \textit{human} dissimilarity judgment, $ED$ the edit distance over the phonetic transcriptions of segments (streches of speech between two time-stamps), $D(U)$ the total duration of dataset $U$ (in sec), $p$ the probability of a given discrete unit $u_i$ in $U$, $\hat{p}$ a pseudo-probability computed by the model over an input sequence (word or sentence). * indicates ungrammaticality}
\end{tabular}
\label{tab:tasks}
\end{table*}

Preschoolers around the world demonstrate clearly that it is possible to learn to naturally interact in language without knowing how to  read or write \cite{dupoux2018cognitive,bavin_cambridge_2009}. Written language is, in a way, a tool for archiving spoken or signed language. Many languages have no writing system at all, and many other language communities do not use the written form of their language much (reportedly, more than half the world's languages do not have a stable or widely used writing system).

Reverse engineering the feat of learning a spoken language from speech input only is a fascinating scientific question. The Zero Resource Speech Challenge (ZRC) focuses on spoken languages. (It is an important and pressing question how zero-resource technology could be applied to working with signed languages.) For spoken language technology, advancing this question would unlock a number of novel applications. For one thing, it would allow for applications that address the needs of languages that are entirely or mostly unwritten. Even in languages with large amounts of textual resources, learning language representations from raw audio would help capture the dimensions of language that are typically poorly represented in text (prosody, emotional and non-verbal vocalizations, and so on). Beyond helping to model these aspects of language, capturing unwritten oral expression could improve the ability of machine learning systems to deal with spontaneous speech, thereby unlocking the rich syntax and vocabulary of oral registers, which very often differ greatly from the written form. This would foster more naturalistic human/machine interactions. 

While some research has focused on how self-supervised modelling can improve existing supervised speech tasks (for example, the SUPERB benchmark: \cite{yang2021superb}), the Zero Resource Challenge series assesses progress toward spoken language systems that function without any textual supervision at any point. Building a text-free dialogue system using only raw speech is a difficult machine learning challenge. It requires us to rethink the spoken language processing stack from the ground up. The ZRC series has been designed to address two interlocking research problems: the task problem and the evaluation problem. 

The \textit{task problem} 
is to break down the ill-defined objective of ``learning to process spoken language  without  text'' into a series of well-defined sub-problems. The ZRC series follows the general architecture of a spoken digital assistant to define the learning problem implied by the training of each component---the acoustic model, a lexicon, language models, waveform generation, and so on. But instead of using phonemes, characters or words as an intermediate representation, the components are allowed to develop their own latent representations. For instance, instead of outputting phonemes or characters, the acoustic model is assumed to output \emph{acoustic units} which may or may not be discrete. Such an architecture (see Figure \ref{fig:blueprint}b) naturally gives rise to the following four tasks: (Task 1) Acoustic Unit Discovery; (Task 2) Spoken Term Discovery; (Task 3) Unsupervised Discrete Resynthesis; (Task 4) Spoken Language Modeling. These are the textless counterparts of well-known tasks: (Task 1) phone recognition; (Task 2) word recognition (i.e., ASR); (Task 3) TTS; (Task 4) Language Modeling. We will review these tasks in turn in the following sections. 

The \textit{evaluation problem} is to define metrics that enable model comparison and cumulative progress for tasks that are defined only through a self-supervised  objective. For instance, ASR systems can readily be measured through phone or word error rates. But their self-supervised counterparts, Acoustic Unit Discovery systems, do not aim at recovering phonemes, but a latent representation. How can we evaluate theses systems?  Interest in some of the above mentionned tasks predates the ZRC series (see for instance \cite{kohonen_1988,varadarajan2008,huijbregts2011,lee2012,jansen_2013,badino2014}, for Task 1), but since each of the published papers used its own metrics, it was difficult (and still is) to compare systems and measure progress. The general strategy of the ZRC series is to develop zero-shot probe tasks that are inspired by human psycholinguistics, and which require no model retraining. The reasoning is that, since the aim is to probe for latent representations at various levels of a self-supervised system, it is best to not train any classifier on top of it. Otherwise, it would be unclear whether the performance obtained reflects the system under observation or simply the classifier. For each task, zero-shot metrics were developed that probe for the different levels of linguistic knowledge that the self-supervised system is supposed to have learned. They only require the extraction of information readily available in the system (for example, embeddings, pseudo-probabilities), and are computed by a separate fixed module which is identical across systems. The evaluation metrics that go with the tasks are listed in Table \ref{tab:tasks} and will be presented in more detail in the following sections.

In this paper, we provide a comprehensive overview of the results obtained across the different tasks and metrics of the ZRC series since 2015, and we discuss the lessons learned  and outline the areas that need more work or give puzzling results.

\begin{table}[]
\setlength{\tabcolsep}{3.9pt}
\centering
\caption{\textbf{Summary of tasks and benchmarks in the Zero Resource Challenge series.}} \label{tab:series}
\begin{tabular}{lll}
\hline
Chall.  & Tasks             & Benchmark\\
\hline
2015\cite{versteegh_2016}   & T1, T2    & ABX-15, TDE-15  \\
2017\cite{dunbar2017zero}   & T1, T2    & ABX-17, TDE-17   \\
2019\cite{dunbar2019zero}   & T3        & TTS0-19\\ 
2020\cite{dunbar2020}       & T1,T2,T3  &  \textit{reboot of 2017 \& 2019} \\
2021a\cite{dunbar2021zero}  & T1,T4     & ABX-LS, sLM-21 (sWUGGY, \\
2021b                       & T1,T4     & \hspace{1.6cm}sBLIMP, sSIMI)\\
\hline
\end{tabular}
\end{table}

\section{Past and present}

Six editions of the Zero Resource Challenge have been proposed over the years as events in different venues (Interspeech, ASRU, NeurIPS) and are summarized in Table \ref{tab:series}. Each edition has explored a different combination of tasks and introduced different datasets. Overall, the six editions have received a total of 115 submissions from 29 teams. In addition, several papers have been published using some of the Zero Resource benchmark metrics, which we also include in our review.

\begin{table}[]
\caption{List of contributed systems for each benchmarks}\label{tab:systems}
\vspace{-0.5em}

\begin{tabular}{lrlll}
\hline
System  & Date & Paper                     & Author affiliation & Model type\\
\hline
\multicolumn{5}{c}{\underline{\it ABX-15 (Task 1)}}\\
Bad15a--c& 2015 &  \cite{badino2015discovering} &  LCSL IIT--MIT & AutoEnc                  \\
Bal15a--e& 2015 &  \cite{baljekar2015using}     &  CMU           & Other          \\
Che15a--d& 2015 &  \cite{chen2015parallel}      &   ASTAR        & GMM    \\
Fah15    & 2015 &  \cite{myrman2017partitioning}&   KTH          & Siamese         \\
Hec16a--d& 2016 &  \cite{heck2016unsupervised}  & NAIST          & GMM \\
Ren15a--d& 2015 &  \cite{renshaw2015comparison} &  Edinburgh     & AutoEnc         \\
Sri15a,b& 2015  &\cite{srivastava2016articulatory}& IIT Hyderabad& AutoEnc        \\
Thi15   & 2015  &  \cite{thiolliere2015hybrid}   & CoML          & Siamese        \\
Zeg15   & 2015  &  \cite{zeghidour2016deep}      & CoML          & Siamese        \\
\hline
\multicolumn{5}{c}{\underline{\it ABX-17 (Task 1)}}\\
Ans17a--d& 2017 &  \cite{ansari2017deep}         & LEAP-IIS     & GMM/AutoEnc    \\  
Che17a,b& 2017  &  \cite{chen2017multilingual}   & NU Singapore & GMM     \\  
Hec17a,b& 2017  &  \cite{heck2017feature}        & NAIST        & GMM    \\  
Pel17a,b& 2017  &  \cite{pellegrini2017technical}& Toulouse     & Other     \\ 
Ras17   & 2017  &  Unpub                         & Aalto        & Other    \\ 
Shi17a,b& 2017  &  \cite{shibata2017composite}   & Tokyo IT     & Other    \\ 
Yua17a-c& 2017  &  \cite{yuan2017extracting}     & NU Singapore &  AutoEnc   \\ 
Cho19   & 2019  &  \cite{chorowski2019unsupervised}& Wrocław/DMind& AutoEnc\\
Kha20   & 2020  &  \cite{kharitonov2021data}     & Meta         & Predict.    \\
\hline
\multicolumn{5}{c}{\underline{\it TDE-15 (Task 2)}}\\
BAS2    & 2015 &  \cite{jansen2011efficient}    & JHU          & Match-first\\ 
Kam15a  & 2015 & \cite{kamper2017embedded}      &  Stellenbosch& Match-first\\ 
Lyz15a-c& 2015 & \cite{lyzinski2015evaluation}  &  JHU         & Match-first \\ 
Ras15a-c& 2015 &  \cite{rasanen2015unsupervised}&  Aalto       & Match-first\\ 
\hline
\multicolumn{5}{c}{\underline{\it TDE-17 (Task 2)}}\\
BAS2    & 2017 &   \cite{jansen2011efficient}   & JHU     & Match-first  \\
Kam22   & 2022 &\cite{kamper2022word}           & Stellenbosch &  Segment-first \\
Alg22   & 2022 &\cite{algayres22dpparse}        & CoML         &   Segment-first \\
Kam17   & 2017 &\cite{kamper2017segmental}      & Stellenbosch &  Segment-first  \\
Bha20a,b& 2020 &  \cite{bhati2020self}          & JHU          & Segment-first  \\
Gar17a,b& 2017 &   Unpub                        & UP València  &  Match-first \\
Ras20a,b& 2020 &   \cite{rasanen2020unsupervised}&     Aalto   &  Match-first  \\
Ras17   & 2017 &   \cite{seshadri2017comparison} & Aalto       & Match-first \\
Iwa21   & 2021 &   \cite{iwamoto2021unsupervised}& Tokyo IT    &  Segment-first \\
\hline
\multicolumn{5}{c}{\underline{\it TTS0-19 (Task 3)}}\\
BAS3    & 2019 &  \cite{DBLP:conf/sltu/OndelBC16}& Brno        & AutoEnc/Low br \\
Tja19a,b& 2019 & \cite{tjandra2019vqvae}        & NAIST        & AutoEnc/High br\\
Hor19a,b& 2019 &   Unpub                         &  Horizon    & AutoEnc/High br\\
Pan19a,b& 2019 &   \cite{murthy2020zero}         & IIT Madras  & Other/Low br\\
Kam19a,b& 2019 &     \cite{eloff2019unsupervised}& Stellenbosch& AutoEnc/High br \\
Fen19a,b& 2019 &   \cite{feng2019}               & CUHK        & AutoEnc/High br\\
Ral19   & 2019 &  Unpub                          & CMU         & GMM/Low br\\
Cho19a,b& 2019 &   Unpub                         & Korea U     & AutoEnv/High br  \\
Liu19a,b& 2019 &    \cite{liu2019unsupervised}   & Nat Taiwan U& AutoEnc/Low br\\
Kum19a,b& 2019 &  \cite{nayak2019virtual}        & IIT Hyderabad&AutoEnc/Low br \\
Yus19   & 2019 &    \cite{yusuf2019temporally}   & Boğaziçi    & Other/Low br\\ 
Gök19   & 2019 &  \cite{yusuf2019temporally}     & Boğaziçi    & AutoEnc/Low br\\
Gün20 & 2020 &   \cite{gundogdu2020vector}       & Boğaziçi    & AutoEnc/Low br\\
Yus20a-c& 2020 &   \cite{yusuf2021hierarchical}  & Brno        & Other/Low br\\ 
Hou20a,b& 2020 &  Unpub                          &  Tokyo IT   & AutoEnc/High br\\
Mor20   & 2020 & \cite{morita2020exploring}      & Kyoto       & AutoEnc/High br \\ 
Kum20a,b& 2020 &  \cite{prakash2020exploration}  & IIT Madras  & Other/Low br  \\ 
Che20a,b& 2020 & \cite{chen2020unsupervised}     & Sheffield   & AutoEnc/High br\\ 
Lum20a,b& 2020 &  \cite{tobing2020cyclic}        & Nagoya      & AutoEnc/High br\\
Nie20a,b& 2020 &  \cite{van2020vector}           & Stellenbosch& Other/High br \\ 
Tja20a,b& 2020 &  \cite{tjandra2020transformer}  & NAIST       & AutoEnc/High br\\
\hline
\multicolumn{5}{c}{\underline{\it ABX-LS, sLM-21 (Task1, Task4)}}\\
BAS4 & 2021 &    \cite{nguyen2020zero}           & CoML        & Predict.+BERT \\
BAS4vg & 2021&   \cite{alishahi2021zr}           & CoML        & Predict.+BERT \\
Nie21    &  2021& \cite{van2021analyzing}        & Stellenbosch& Predict.+LSTM\\
Cho21a,b & 2021 & \cite{chorowski2021information}& Wrocław    & Predict./Other \\
Liu21    &  2021& Unpub                          & PLA IEU    & Predict.+LSTM\\
Mae21a--d& 2021 & \cite{maekaku2021speech}       & Yahoo/CMU  & Predict.+BERT\\
Pen21    & 2021 &  \cite{peng2022self}           &  UT Austin & Predict.+BERT \\
Lee21a,b & 2021 & Unpub                          & Seoul      & Other+BERT \\
Gan21    & 2021 & Unpub                          & LEAP-IIS   & Predict.+LSTM \\
Ngu21a--d& 2021 & \cite{nguyen22discrete}        &  CoML      & Predict.+BERT\\
Łań21    & 2021 & Unpub                          &  Wrocław   &Predict.+LSTM  \\
Bha21a   & 2021 & \cite{bhati2021unsupervised}   &  JHU       & Predict. \\
Gao21a--c& 2021 & Unpub                          &  TU Delft  & Predict.+BERT \\
Qia22 & 2022 & \cite{qian2022contentvec}         & MIT--IBM   & Predict.+Transf \\
Cue22 & 2022 & \cite{cuervo2021contrastive}      & Wrocław    & Predict.\\
\hline
\end{tabular}
$\,$\vspace{0.2em}$\,$
\textbf{Note.} br: bitrate. See model type details in Section II. 
\end{table}

Table \ref{tab:systems} gives the complete list of submitted systems to all four tasks, and the abbreviations we use for them, including citations for published systems and model types as explained in the upcoming sections.

\begin{table*}[t]
\centering
\def\arraystretch{1.2} 
\setlength{\tabcolsep}{3.0pt}
\caption{\textbf{Characteristics of the ZRC benchmark datasets. $^*$for TDE-17, test set is same as train set.} \label{tab:zrc_datasets}}
\begin{tabular}{llllp{4cm}}
\hline
\textbf{Benchmark}& \textbf{Language (source dataset)}                                                      & \textbf{Type}                 & \textbf{(Default) Train Set} & \textbf{Test Set}  \\
\hline
ABX-15, TDE-15    & \begin{tabular}[c]{@{}l@{}}English (Buckeye)\\ Xitsonga (NCHLT)\end{tabular}            
          & \begin{tabular}[c]{@{}l@{}}conversations\\ TIMIT-like\end{tabular}  
          & Same as test  
          & \begin{tabular}[c]{@{}l@{}}5h, 12 spk\\ 2h30, 24 spk\end{tabular} \\ \cline{2-5}

ABX-17, TDE-17*   & \begin{tabular}[c]{@{}l@{}}English (Librivox)\\ French (Librivox)\\ Mandarin (THCHS-30) \\ German[L1] (Librivox)\\ Wolof[L2] (\cite{gauthier:hal-01350037})\end{tabular} 
          & \begin{tabular}[c]{@{}l@{}}audiobook\\ audiobook\\ read speech\\ audiobook\\ TIMIT-like\end{tabular} 
          & \begin{tabular}[c]{@{}l@{}}45h\\ 24h\\ 2h30\\ 25h\\ 10h\end{tabular}  
          & \begin{tabular}[c]{@{}l@{}}27h, 9 spk\\ 17h, 10 spk\\ 25h, 4 spk \\ 11h, 10 spk\\ 5h50, 4 spk\end{tabular} \\ \cline{2-5}
          
ABX-LS            & English (LibriSpeech)  & audiobook  & 960h or 100h  & \begin{tabular}[c]{@{}l@{}}dev/test x clean/other: 5h each\\ 40 spk for clean,\\ 33 spk for other\end{tabular}  \\ \cline{2-5}

TTS0-19           & \begin{tabular}[c]{@{}l@{}}English (Librispeech)\\ Indonesian (\cite{sakti2008developmentasr,sakti2008developmentsynth})\end{tabular} & \begin{tabular}[c]{@{}l@{}}audiobooks\\TIMIT-like \end{tabular}  &\begin{tabular}[c]{@{}l@{}}Voice: 4h40, 2 spk; Unit: 15h, 100 spk\\Voice: 1h30, 1 spk; Unit: 15h, 112 spk\end{tabular} & \begin{tabular}[c]{@{}l@{}}28min, 24spk\\29 min, 15 spk\end{tabular} \\ \cline{2-5}
\begin{tabular}[c]{@{}l@{}}sLM-21 (sWUGGY,\\sBLIMP, sSIMI)\end{tabular} & English (LibriSpeech)& audiobook & 960h&
\begin{tabular}[c]{@{}l@{}}Synthetic words or short sentences\\or extracted from LibriSpeech\end{tabular}  \\ 
\hline
\end{tabular}
\end{table*}

\subsection{Task 1: Acoustic Unit Discovery}
\label{sec:t1}


The goal of acoustic unit discovery is to learn representations (embeddings) of speech sounds that retain linguistically relevant information 
and discard acoustic information which is irrelevant or secondary to recovering the linguistic content, like speaker voice type or recording conditions (additive noise, reverberation, etc). In text-based systems, such representations are typically phonemes (as defined by a pronunciation dictionary) or characters. Here, we let the representations be latent, which means that they may take any form (dense vectors for each frame, probabilistic codes, discrete codes, etc). This poses an evaluation problem. The ZRC series takes the view that, while discovered units may not necessarily take the shape of straighforwardly linguistically interpretable entities like phonemes or phonetic features, they should at least maintain the same key linguistic function: \emph{linguistic contrast}.

\subsubsection{Evaluation metrics}

Phoneme categories are typically defined as the smallest element of speech that can induce a  difference in meaning between words. In English, for instance, the phonemic contrast between /r/ and /l/ is demonstrated by the fact that ``fly'' and ``fry'' are distinct words.  Two instances of ``fly'' would remain instances of the word ``fly'' even in spite of  variations in speaker or recording condition, and an instance of ``fry'' is (for speakers with a standard pronunciation) never an instance of ``fly.'' The same goes for possible, as opposed to actual, words: ``pla'' and ``pra'' would not be the same word, if they were words.

The minimal pair \textbf{ABX} task \cite{schatz2013,schatz2014}  is inspired by match-to-sample tasks used in human psychophysics and measures discriminability between two sound categories. 
We define ${\Delta}$, the ABX-discriminability of category $A$ from category $B$, as the probability that tokens $a$ and $x$ from $A$ are further apart than token $b$ from $B$ and $x$ are, according to a dissimilarity function $d$, see Equation \ref{eq:abx}:
\begin{multline}\small
{\Delta}(A,B) ={ \frac{1}{|A|(|A|-1)|B|} \sum_{a\in A} \sum_{b\in B} \sum_{x\in A \setminus {a}} (\mathds{1}_{d(a,x)<d(b,x)} } \\+ \frac{1}{2}\mathds{1}_{d(a,x)=d(b,x)})     
\label{eq:abx}
\end{multline}
where $\mathds{1}$ is the indicator function and $|A|$ the number of tokens in category $A$. The discriminability score is symmetrized by averaging $\Delta(A,B)$ and $\Delta(B,A)$. 

Tokens are representation of speech sequences as output by the model under evaluation.
In general, they will be sequences of frame embeddings, and dynamic time warping is used to realign them. Frame-level dissimilarities are averaged along the realignment path to obtain $d$. Most submissions used angular dissimilarity (arccos of the normalized dot product of the frame embeddings), while others used the KL divergence when the frame representations were posterior probabilities. 

For the categories $A$ and $B$ we use triphone sequences that only vary in the middle phoneme (like ``fly'' and ``fry''), as extracted from longer utterances in our test set. Thus, participants apply their trained models to these audio files, and output sequences of representations for the entire file. These representations must be time-stamped, so that the ABX evaluation software can identify the beginning and end of each three-phone token in the output. This was done in all three \textbf{ABX-} benchmarks; only the ABX evaluation that was contained in the \textbf{TTS0-19} evaluation package did otherwise, using small audio files containing only the three-phone sequence (see Section \ref{sec:task3})\footnote{The risk of the time stamping approach is that some models' representations may warp time in unpredictable ways, resulting in apparent errors due to misalignment with the time-stamps presupposed by the evaluation. On the other hand, providing very short audio clips at test time may cause its own problems if models have been trained with longer sequences.}. 


For the
\textit{within-speaker} variant of this task, all of the phone triplets belong to the same speaker (e.g., $a=\textrm{fly}_{T_1}$, $b=\textrm{fry}_{T_1}$, $x=\textrm{fly}'_{T_1}$). The error rates for a given minimal  pair  are first averaged across all of the speakers for which this minimal pair exists. The resulting scores are then averaged over all found contexts for a given pair of central phones (e.g., for the pair /l/-/r/, average the scores for all attested contexts such as /f\_y/, /f\_i/, /a\_o/, etc.). Finally, the scores for every pair of central phones are averaged and subtracted from 1 to yield the reported within-talker \textbf{ABX error rate.} For the \textit{across-speaker} variant, $a$ and $b$ belong to the same speaker, and $x$ to a different one. $a=\textrm{fly}_{T_1}$, $b=\textrm{fry}_{T_1}$, $x=\textrm{fly}_{T_2}$. The scores for a given minimal pair are first averaged across all of the pairs of speakers for which this contrast can be made. As above, the resulting scores are averaged over all contexts over all pairs of central phones and converted to an error rate. 

\subsubsection{Datasets} As seen in Table \ref{tab:zrc_datasets}, the first two benchmarks (ABX-15 and ABX-17) used relatively modest datasets sizes for training (from 2.5h to 45h) over 6 languages. ABX-15 used the training set as test set, while the ABX-17 introduced a separate test set with new speakers, to test for the ability of the learned representations to generalize to new speakers. The more recent benchmark (ABX-LS) uses the full LibriSpeech (960h) as training set, and introduced a separate dev and test set in order to avoid overfitting the model's hyperparameters to the test set. As more recent models tend to use more and more data, the benchmarks are open to submissions of systems trained on datasets other than the default ones, so long as they contain no labels and are described in detail.  

\subsubsection{Baselines}
For Task 1, our low end reference score (``baseline'') is calculated by computing the distances on spectral representations (MFCC). Good representations that highlight linguistically relevant differences, and thus neutralize speaker or channel differences, should at least do better than MFCCs. On the high end, using the gold annotations to generate a frame-by-frame one-hot phonemic representation mechanically leads to a perfect ABX score. To give a fairer high end comparison, we have also often included scores for the output of an off-the-shelf supervised GMM--HMM ASR. We included such a ``topline'' score in the \textbf{ABX-15} and \textbf{ABX-17} results, as well as in the unit evaluation component of the \textbf{TTS0-19} evaluation, which also contained an ABX test (see below). In the first two cases, the representations we used were posteriorgrams (that is, rather than a one-hot vector at each frame representing the decoding, we calculate the model's posterior decoding probability for each frame). This reference score was beaten not long after the 2017 challenge (by \textbf{Cho19}), and remains quite poor compared to modern systems (see below).  In the \textbf{TTS0-19} evaluation, we instead used the hard decoding, rather than probabilities, and observed that this reference score was in fact \emph{very} easy to beat (for one of the two languages, the MFCC scores were already better), because of the numerous errors in decoding made by the off-the-shelf ASR system. More recently, submitted systems have become so good at the ABX task (see Figure \ref{fig:task1} and below) that such low-fidelity topline systems have become less relevant. 


\newcommand{\set}[1]{\left\{#1\right\}}
\newcommand{\tup}[1]{\langle#1\rangle}
\newcommand{\cdisc}{C_{\textrm{disc}}}
\newcommand{\pdisc}{P_{\textrm{disc}}}
\newcommand{\fdisc}{F_{\textrm{disc}}}
\newcommand{\fall}{F_{\textrm{all}}}
\newcommand{\pall}{P_{\textrm{all}}}
\newcommand{\pgold}{P_{\textrm{all}}}
\newcommand{\psubs}{P_{\textrm{disc*}}}
\newcommand{\pclus}{P_{\textrm{clus}}}
\newcommand{\bdisc}{B_{\textrm{disc}}}
\newcommand{\bgold}{B_{\textrm{gold}}}
\newcommand{\pgoldclus}{P_{\textrm{goldcl}}}
\newcommand{\pgoldlex}{P_{\textrm{goldLex}}}
\newcommand{\fgoldlex}{F_{\textrm{goldLex}}}
\newcommand\SetB[3]{\ensuremath{\{\text{\ensuremath{#1 \mid} \parbox[t] {\widthof{\ensuremath{#3}}} {\ensuremath{#2}\}}}}}
\newcommand\SetA[2]{\ensuremath{\{\text{\ensuremath{#1 \mid} \parbox[t] {\widthof{\ensuremath{#2}}} {\ensuremath{#2}\}}}}}
\newcommand{\cover}[1]{\mathrm{cover}(#1)}
\newcommand{\flatten}[1]{\mathrm{flat}(#1)}
\newcommand{\nmatch}[1]{\mathrm{occ}(#1)} 
\newcommand{\weight}[1]{\mathrm{weight}(#1)}
\newcommand{\types}[1]{\mathrm{types}(#1)}
\newcommand{\freq}[1]{\mathrm{freq}(#1)}
\newcommand{\ned}[1]{\mathrm{ned}(#1)}
\newcommand{\card}[1]{\left\vert{#1}\right\vert}



\subsubsection{Results}

\begin{figure}[]
\centering
\includegraphics[width=1\columnwidth]{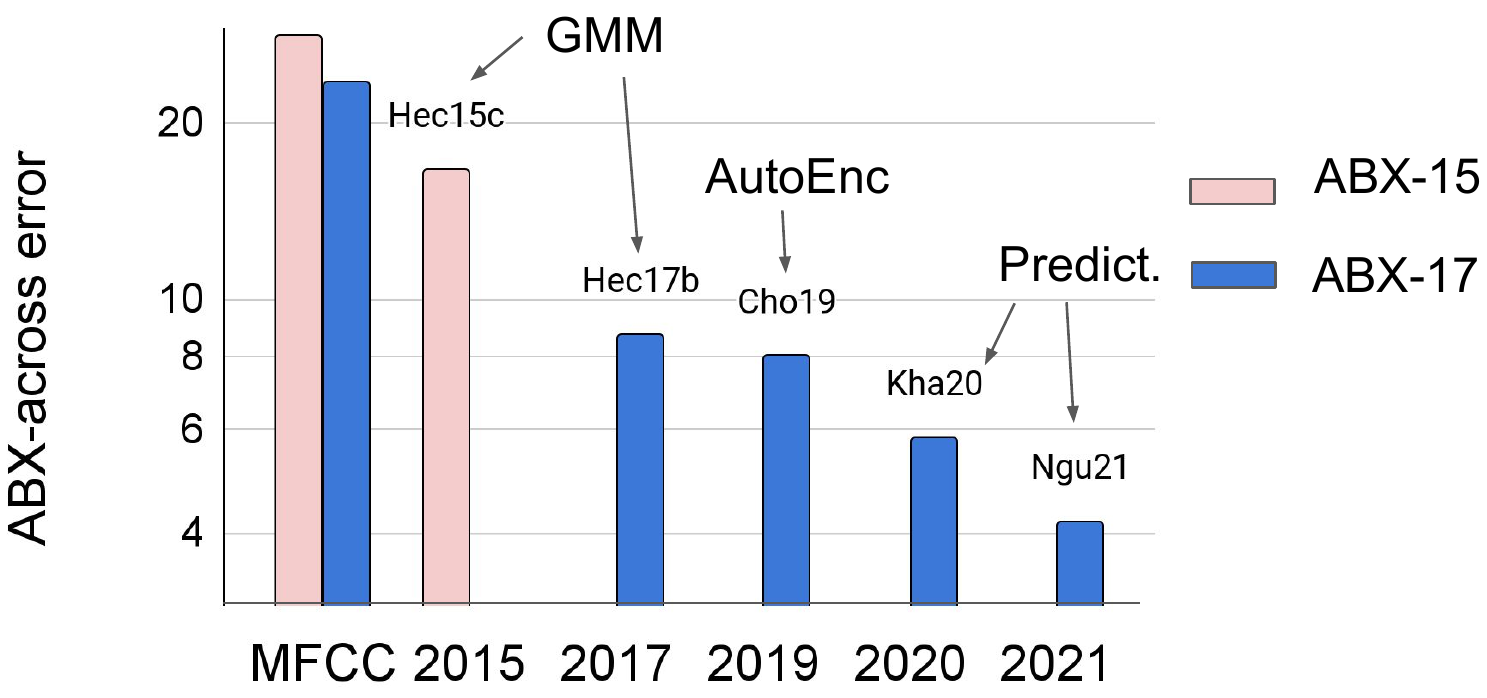}
\caption{ZR Task 1 results on English ABX test sets (ABX-15: Conversational speech--Buckeye; ABX-17: Audiobooks--LibriVox). The left two scores are on MFCC representations. The right two scores have been trained on Librispeech 960. See Table \ref{tab:systems} for references.}
\label{fig:task1}
\end{figure}

Since 2015, several approaches have been taken to Task 1. Most start from the principle that a central characteristic of text (or phonemic transcription) is that it provides a highly \textit{compressed latent representation} for speech. For reference, a 16~kHz 16-bit waveform is coded using 256 Kb/sec, which generic audio codecs like Opus or MP3 can compress down to between 32 and 16~Kb/sec (a factor of 8 to 16). In contrast, a phonemic transcription is about 70 b/sec. This represents a compression of more than 200x compared to general audio codecs! 
Many objective functions proposed for Task 1 have as their primary goal to reduce the amount of information coded.  

A simple and remarkably successful version of this idea---inspired by the \textit{universal background models} used in speaker encodings---is to model acoustic frames using a mixture of full-covariance Gaussians (\textbf{GMM}). The posteriorgram of the mixture is taken as a new, sparse code for the speech input. In other words, each acoustic frame in the input file is assigned a sparse vector of probabilities, which correspond to the a posteriori probability of each of the discovered Gaussian distributions as the source of the given frame. Since individual frames are very short, and they are clustered as independent observations, this gives rise to a code which classifies speech in terms of short-term acoustic events, typically with several hundred Gaussian clusters discovered. This strategy, supplemented with additional speaker normalization or teacher-student tricks, was able to obtain top scores in the 2015 and 2017 editions (\textbf{Che15a\nobreakdash--d, Hec17a,b}: see Figure \ref{fig:task1}).

Another type of approach seeking to find a compressed latent representation uses \textit{autoencoders} (\textbf{AutoEnc}), which aim at reconstrucing the signal through an information bottleneck, sometimes achieved by using a discrete codebook: \textbf{Bad15a\nobreakdash--c, Cho19}.
The codebook~+~WaveNet autoencoder of \textbf{Cho19} obtained better results than previous, mixture model-based systems.

Since 2020,  a new generation of \textit{predictive models} (\textbf{Predict}) began to appear which have given rise to never-before-seen performance: contrastive predictive coding, or CPC \cite{oord2018cpc}, wav2vec 2.0 \cite{baevski2020wav2vec}, HuBERT \cite{hsu2021hubert}; see \cite{borgholt2022brief} for a review. Two salient differences with this new wave of models is how they integrate context, and how they scale by working from the raw audio.  The compression based models tended to have a frame-based view of the speech signal, modeling the probability distribution of individual acoustic frames through a compressed latent representation. In contrast, the predictive models aim to reconstruct large missing or masked parts of the signal \emph{conditional} on visible parts of the sequence. For instance, CPC predicts future frames within a 10 to 120 ms window based on past frames, and obtained excellent ABX scores (\textbf{Kha20, Ngu21, BAS4-sm,lg}: around 4.5\% across speaker). See a more thorough discussion of CPC in the section about Task 4 below. Wav2vec 2.0 and HuBERT try to reconstruct a masked part of the signal (of the order of 100ms), based on left and right context.

Independently of their predictive objectives, these models are also more sophisticated than previous ones in their encoding of context. Instead of processing signals within small receptive fields (\textbf{Bad15a--c,Thi15,Che15a--d}), the new systems use recurrences, transformers, or both, allowing them to model temporal correlation at longer distances. At the phonological level, language can be viewed as an autocorrective code due to redundancies introduced by phonotactics and lexical regularities. Previous work showed that top-down lexical context \cite{feldmanlexical,thiolliere2015hybrid} or phonotactics \cite{elsner2012bootstrapping,ddi} can indeed help with the discovery of phonetic units. This may explain why predictive/masked objectives together with large receptive fields help learning the acoustic properties of units jointly with their functional role in the language, yielding more relevant units.


 The new models are also large, and, accordingly, are typically trained on large amounts of data (thousands of hours), which is orders of magnitude more than the training sets used in the initial ZRC series. In addition, some of the new models work \emph{directly from waveform} instead of relying on engineered features like MFCC or mel filterbanks.  Allowing models to be large, working from the raw audio, and training them on large amounts of data might push them to mimic the evolution of the human auditory system and its  adaptations to speech. Indeed, \cite{millet2021predicting,millet2022self} find that wav2vec 2.0, HuBERT, and, particularly, CPC, are good predictors of low-level (sub-phonemic) auditory and speech processing in humans. In addition, it is well known that human perception relies on temporal fine structure not captured by magnitude spectrograms  \cite{moore2012introduction}, particularly in difficult listening conditions. Models working from waveform might have an advantage.\footnote{We note, however, that, as \cite{weerts2021psychometrics} demonstrates, wav2vec 2.0's use of temporal fine structure still differs from that of human listeners.}

To sum up, predictive models seem to have an edge on compressive models, and enjoy increasing popularity for a variety of downstream applications (see models presented in the SUPERB benchmarks \cite{yang2021superb,tsai2022superb}). In the context of the ZRC series, a fair comparison equating architecture and dataset size would be necessary before claiming a definitive win. In addition, new models combining the two ideas like Masked AutoEncoders are emerging and need to be tested \cite{huang2022MAE,baade2022MAE}. 

Other interesting ideas have been explored in the ZRC series. Although they have not made it to the top of the leaderboard, they may still have much to contribute, perhaps in combination with other approaches. 
For example, some systems have attempted to use possible synergies with Task\,2, and have used Spoken Term Discovery to obtain pairs of segments that have potentially the same phonemes, and use them in a \textbf{Siamese} architecture. Such pairs have also been used as a form of data augmentation with some \textbf{AutoEnc} models as well. In addition,  
most systems have not attempted to model the duration of linguistic units like phonemes or syllables yielding representation of much shorter duration acoustic events (10~ms or so). Yet duration is a principal concern of the HMM-based unit discovery system of \textbf{Baseline3} as well as the segmental CPC approach of \textbf{Bha21a}. The approach of \textbf{Pan19a\nobreakdash--b,Kum20a\nobreakdash--b} implicitly considers duration by dividing the signal into syllables, which are then further divided into subsyllabic units (see  also Task 4 in Section \ref{sec:t4}). None of these other approaches has reached state-of-the-art performance yet, but, once again, duration is quite clearly critical to speech perception, and so it seems likely that research will need to examine these ideas further.

\subsection{Task 2: Spoken Term Discovery}\label{sec:t2}


Just as the infant learns the words of their language by listening, Spoken Term Discovery seeks to find recurring patterns in the input, proposing a set of boundaries delimiting the start and end of proposed word tokens discovered in the speech, and category labels indicating proposed word types.\footnote{Note here, that contrary to Task 1, we ask systems to explicitly return linguistically interpretable representations (boundaries and cluster labels) instead of a representation that simply satisfies certain functional properties.} This problem was explored by several papers prior to the ZRC \cite{park2008,zhang2010,jansen2011,muscariello2012,jansen2013}, and served as inspiration for the challenge itself. Although the task of ``finding words'' seems intuitively simple, it is made up of at least three subproblems which we evaluate separately.

\begin{itemize}
\item
The \emph{matching} subproblem is to find all pairs of speech fragments that are instances of the same sequence of phonemes. This can be evaluated based on how phonemically similar the discovered fragments are using the gold transcription (normalized edit distance: \textbf{NED}) and how much of the corpus they cover (\textbf{coverage}).

\item The \emph{lexicon discovery} subproblem is to group these fragments into clusters (as opposed to simple pairwise matching). The goal is to find a lexicon of types. A proposed cluster can be evaluated based on how well the members match on the sequence of phonemes (\textbf{Grouping}) and how well the sets match the gold-standard lexicon of word types (\textbf{Type $F$-score}).

\item The \emph{word segmentation} subproblem attempts to find onsets and offsets of fragments that are aligned with the word boundaries as defined in the gold-standard text.
\end{itemize}

\subsubsection{Evaluation metrics}

To maximize  comparability with text-based word discovery approaches, all of these evaluations are done by forced aligning the test set with its phoneme transcription. Any discovered speech fragment is converted into its transcription (which means taking decisions about phonemes on the left or right edge which may be partially covered: we include a phoneme if the fragment contains more than more than 30~ms of that phoneme or more than 50\% of its duration).

The evaluation of spoken term discovery systems as matching systems consists of two scores,  \textbf{NED} (normalized edit distance)  and  \textbf{coverage}. \textbf{NED} is the average, over all matched pairs, of the Levenshtein distance between their phonemic transcriptions, divided by the max of their phonemic length ($ED(a,b)/max(|a|,|b|)$, where $a$ and $b$ are the two elements of a proposed match). The \textbf{coverage} is the fraction of the discoverable part of the corpus that is covered by the set of all discovered fragments. The discoverable part of the corpus is found by computing the union of all of the intervals corresponding to all of the pairs of n-grams (with n between 3 and 20). This is almost all of the corpus, except for unigram and bigram hapaxes.

Six scores are used to evaluate the performance of a spoken term discovery system in terms of lexicon discovery. The first three are \textbf{grouping precision, grouping recall,} and \textbf{grouping F-score.} These are defined in terms of $\pclus$, the set of all pairs of fragments that are groups in the same cluster, and $\pgoldclus$, the set of all non-overlapping pairs of fragments which are both discovered by the system (not necessarily in the same cluster) and have exactly the same gold transcription. 
\begin{align}
 \mbox{Prec:} & \sum_{t\in\types{{\pclus}}} w(t,\pclus)\frac{\card{\nmatch{t, \pclus\cap\pgoldclus}}} {\card{\nmatch{t,\pclus}}} \label{eq:grouping:precision} \\
 \mbox{Rec:} & \sum_{t\in\types{{\pgoldclus}}} w(t,\pgoldclus)\frac{\card{\nmatch{t, \pclus\cap\pgoldclus}}} {\card{\nmatch{t, \pgoldclus}}}\label{eq:grouping:recall}
\end{align}
where $t$ ranges over the types of fragments (defined by the transcription) in a cluster $C$, and $occ(t,C)$ is the number of occurrences of that type, $w$ the number of occurrences divided by the size of the cluster. In other words, Prec is a weighted measure of cluster purity and Rec, of the inverse of the cluster's fragmentation. The other three scores are \textbf{type precision, type      recall,} and \textbf{type F-score}. Type precision is the probability that discovered types belong to the gold set of types (real words), whereas type recall is the probability that gold types are discovered. We restrict both sets to words between three and twenty phonemes long.

To evaluate systems on the word segmentation subproblem,  we use the \textbf{token} and \textbf{boundary} $F$-scores with respect to the gold text as is usual in text-based word segmentation. The token $F$-score evaluates whether the set of discovered tokens matches the gold set of tokens, and the boundary $F$-score evaluates whether the set of boundaries (the delimitations between tokens, in terms of which phones in the gold transcription are separated by a boundary and which are not) corresponds to the gold set of boundaries.

The baseline system we provide is that of \cite{jansen2011efficient}, which matches acoustic pairs using locally-sensitive hashing and then groups the pairs together using graph clustering. The topline was based on applying the adaptor grammar segmentation algorithm of \cite{adaptorgrammar} to the gold textual transcriptions.

\begin{figure}[]
\centering
\includegraphics[width=\columnwidth]{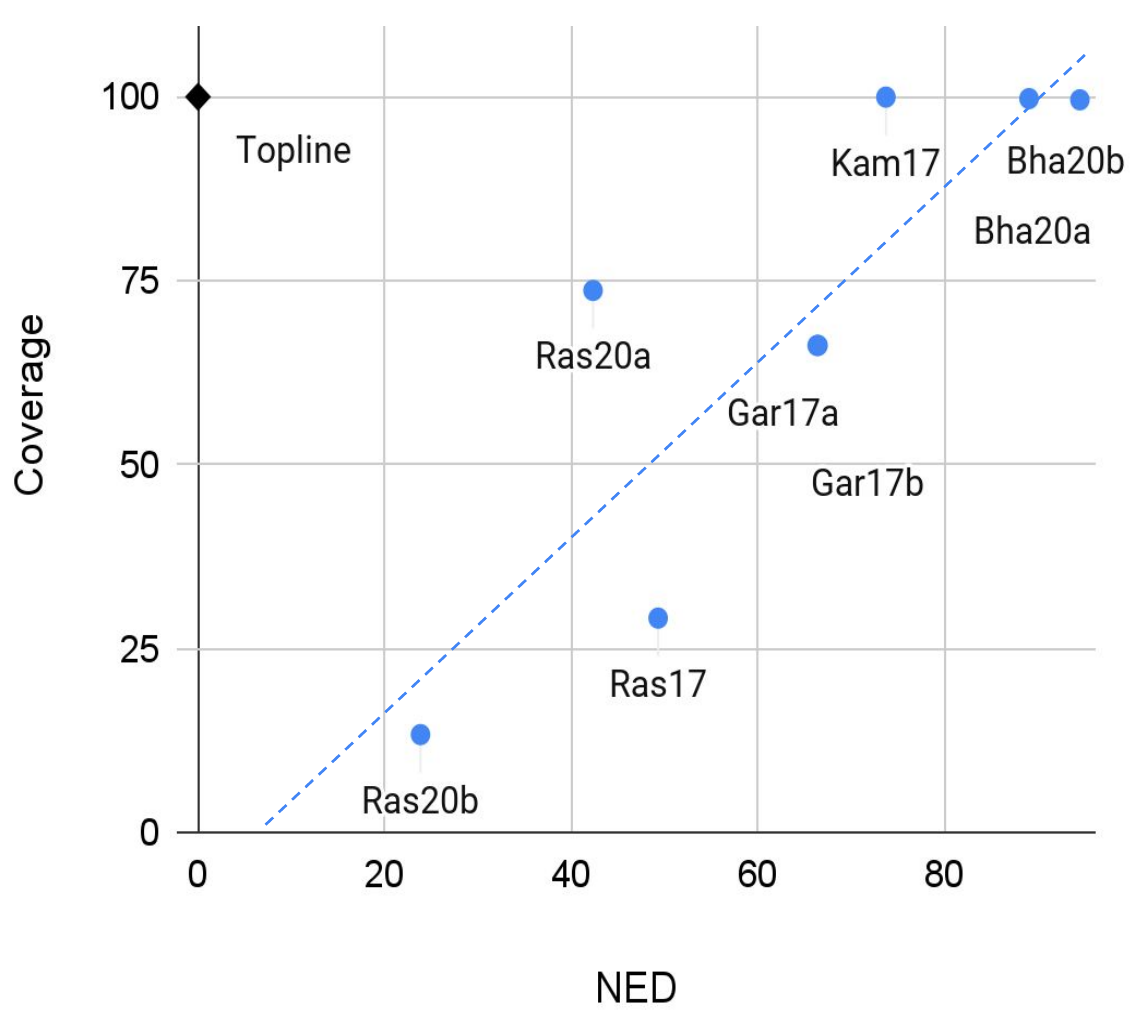}
\caption{Task 3 (term discovery): NED vs Coverage Tradeoff. Average results across 5 languages (ZR17 and ZR20).}
\label{fig:nedcov}
\end{figure}

\subsubsection{Datasets} The datasets for the 2015 and 2017 Task 2 benchmark (TDE-15 and TDE-17, see Table \ref{tab:zrc_datasets}) are coextensive with those of the corresponding Task 1 benchmarks, with one exception: the test set is always same as the training set. This may seem unorthodox from a machine learning point of view, but is quite common in the text segmentation litterature, as the models are evaluated on their ability to extract words and boundaries from the training set. 

\subsubsection{Results}
The Spoken Term Discovery task is still very challenging and has not  received the same attention as Acoustic Unit Discovery. One major finding across the three ZRC editions that featured this task is the existence of a  \textbf{tradeoff} between  attempting to find a lot of words and ensuring that the discovered words are accurate. The quality of the set of words that are treated as matches/repetitions by the system, as measured by the normalized edit distance (NED), will necessarily be better if systems do not commit to extracting more dubious word candidates in the first place; however, the more candidates are ignored, the less of the corpus will receive an analysis (lower coverage) and the fewer of the gold word boundaries will be found (leading also to lower boundary $F$-scores). The tradeoff between term quality and coverage is shown in Figure \ref{fig:nedcov}.

Systems that take a ``matching first'' approach, like \textbf{Ras15a\nobreakdash--c, Ras20a--b,} seek primarily to find recurrent phonetic patterns.  Boundaries here are merely designations of the edges of these discovered segments.  The system that currently does best at balancing NED with coverage and segmentation quality, \textbf{Ras20a}, takes a  matching-based approach, based only on MFCC inputs. This system begins by doing a low-resolution search for candidate matches by dividing the input utterances into fixed-length down-sampled speech segments. Then, it filters the candidate matches using a higher-resolution matching algorithm based on dynamic time warping.

It might seem surprising that an algorithm that uses MFCC inputs, rather than features learned by acoustic unit discovery, would yield good performance.  Nevertheless, \cite{algayres2020evaluating} demonstrated that ABX error rate may not be the best indicator of downstream lexicon discovery, and our own informal experiments have shown that naively feeding improved acoustic units into a generic matching system (for example, those learned by \textbf{Thi15}) can actually make matching quality \emph{worse.}


Other Task 2 systems take segmentation-oriented approaches, putting a priority on discovering boundaries. Building on earlier text-based systems using non-parametric Bayesian models like \cite{johnsonetal2007,adaptorgrammar,goldwater}, systems like  \textbf{Kam15,Kam17} jointly optimize an exhaustive segmentation and a dictionary of clustered word embeddings. In a different, bottom-up approach, \textbf{Bha20a,b} matches learned segmental acoustic units to construct a full word segmentation.  As would be expected, since these systems strive to optimize segmentation measures, they fare rather poorly on matching measures. 

\begin{figure}[]
\centering
\includegraphics[width=.9\columnwidth]{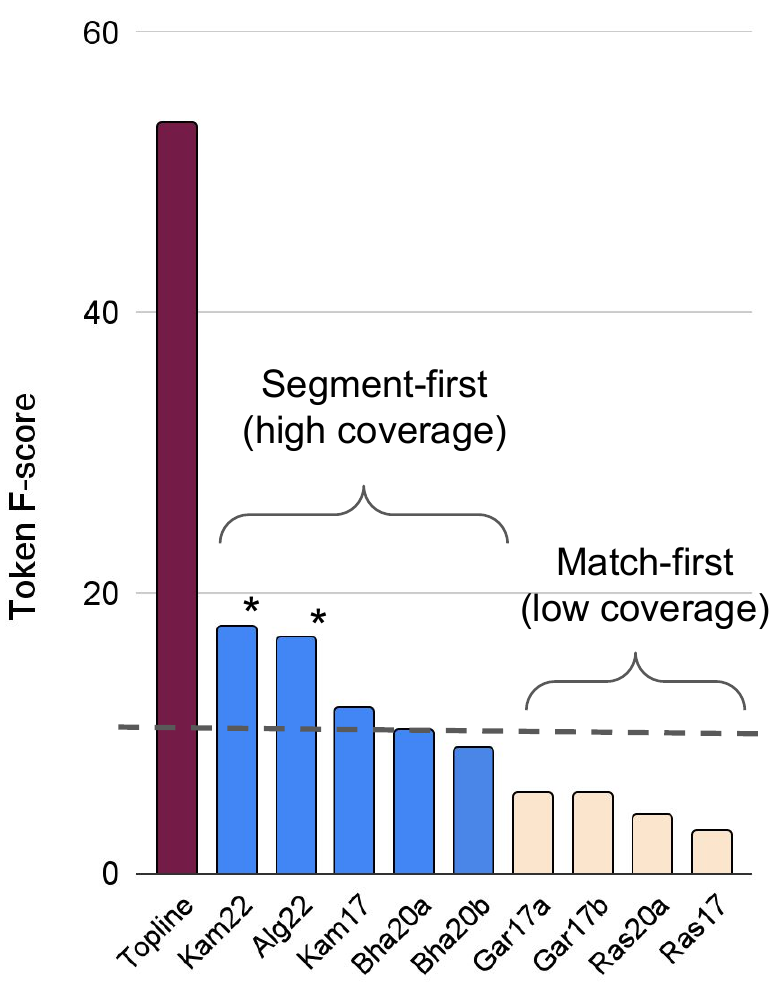}
\caption{Task 2 (term discovery): Token F-scores, measuring how many words are correctly segmented, averaged across 5 languages (ZR17 and ZR20 plus two new papers). The topline is a unigram word segmentation adaptor grammar trained on the same amount of text. The dotted line is a baseline consisting in random segmentations every 120ms. Starred (*) models compute the segmentation without even building any lexicon of discrete types.}
\label{fig:tokenFscore}
\end{figure}

Figure \ref{fig:tokenFscore} focuses on  segmentation by itself and displays the Token F-score for each of the submitted systems, compared to a topline unigram adaptor grammar segmentation system trained on the corresponding text (phonemized text without the blank spaces between words).  All of the high-coverage segmentation-oriented models are on the left and all of the low NED, matching-first models on the  right. The segmentation-oriented models are more likely to do well on this metric, which assesses how many of the true word tokens were correctly segmented. Included here are two new models, \textbf{Kam22} and \textbf{Alg22,} which do not even attempt to build a lexicon of types. The first one is in the same vein as \cite{bhati2021segmental}, which posits word boundaries at peaks in surprisal across sequences of learned segmental units, while the second uses a non-parametric Bayesian approach directly on tokens. As can be seen, however, the gap between the best speech based models and the text-based ones is still large.

The reason is likely multi-fold. First, speech is variable, which means that the same word will surface with a variety of acoustic shapes. Even if good invariant quantized acoustic representations are used, the potential (and actual) variability in the different ``transcriptions'' in terms of these quantized units for the ``same'' word  grows exponentially with word length. This makes it difficult to build a reliable lexicon of word types. Second, speech rate and phone durations are variable in time, with the result that both phoneme duration and word duration can change substantially from occurrence to occurrence, a problem that does not exist in text. Finally, speech is typically coded into frames, which gives it a finer granularity than text (e.g., 10~ms frames, whereas phonemes last on average around 70~ms): a potential word boundary can therefore occur in more places in speech than in text. This increases the number of potential segmentation errors that can be made. This last point is one of the motivations for systems such as \textbf{Kam22, Bha21a, Cue22,} all of which, jointly or sequentially, infer word boundaries hierarchically on the basis of learned acoustic unit boundaries. Any future work will need to address all three of these challenges to achieve better performance on this task. 

\subsection{Task 3: Discrete Resynthesis (TTS without T)}
\label{sec:task3}

\begin{figure*}[]
\centering
\includegraphics[width=2\columnwidth]{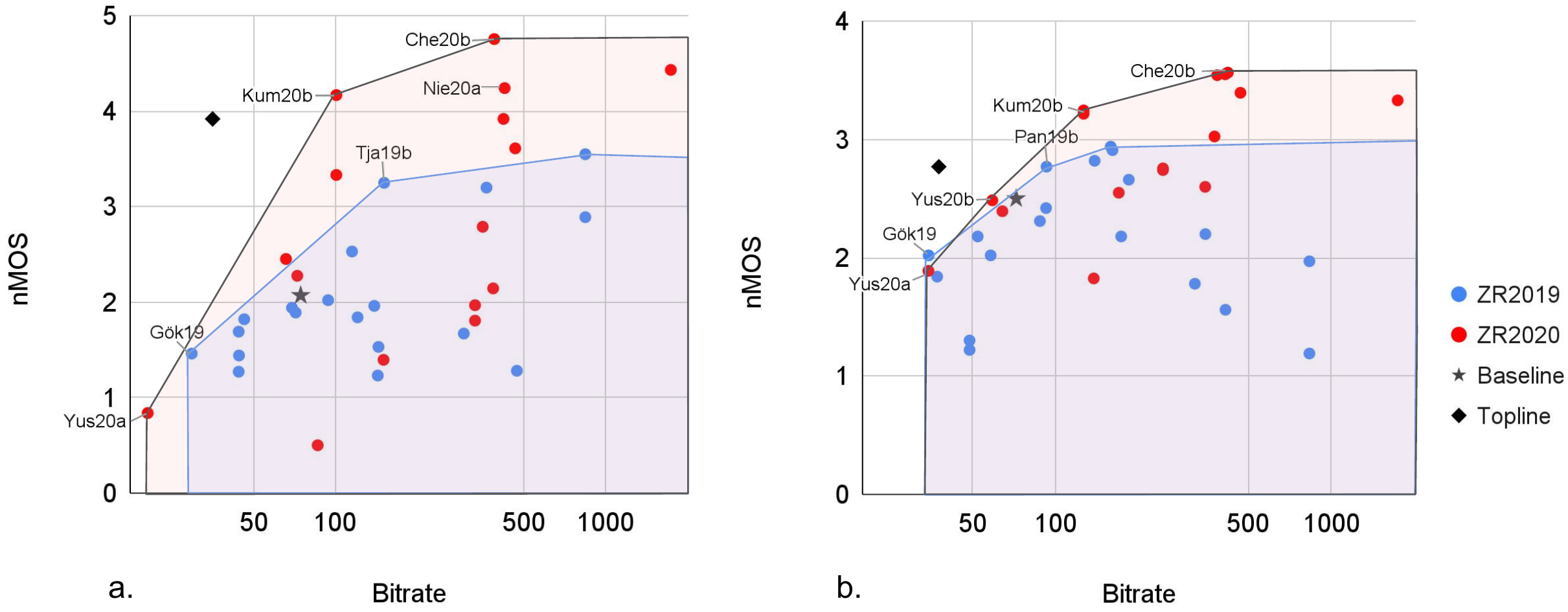}
\caption{ZR Task 3 (TTS without T) human evaluation results on the surprise language (Indonesian, panel a) and the dev language (English, panel b) across the two editions of the Challenge (2019 and 2020). The nMOS score is obtained by normalizing the Mean Opinion Score across the two challenges by using the Baseline and Topline as anchor points. }
\label{fig:MOSbitrate}
\end{figure*}

Here, we investigate a task which is similar to what infants may do when they repeat a word or a sentence: they encode the signal into some representation, and then reproduce the same content in their own voice. Defined like this, the task is already known as voice cloning or voice transfer, and it can be performed at a rather low level by introducing a target speaker embedding in the decoder part of a simple encoder--decoder architecture. Here, however, we add the constraint that there be a discrete bottleneck between the encoder and decoder, and we measure the bitrate of the encoding. In other words, we ask  participants to use discovered acoustic units instead of phonemes, and we push these units to approach the bitrate of phonemic transcription.
Prior to the ZRC, \cite{DBLP:conf/icassp/ScharenborgBBHM18}  demonstrated the feasibility of unsupervised discrete resynthesis. Furthermore,  
some of the models in Task 1 (\textbf{Bad15a\nobreakdash--c,Cho19}) already used a similar discrete bottleneck autoencoder architecture, although they did not evaluate the quality of the reconstruction nor the bitrate of the representation. 
Participants on this task are provided with a \textit{unit dataset} from multiple speakers used to discover discrete units, and a \textit{voice dataset} to train a synthesizer for the target voice taking the units as input. The \textit{test dataset} consists of novel utterances by unseen speakers, which must be resynthesized in the target voice. Participants submit both the acoustic unit representation and the resynthesis for evaluation.

\subsubsection{Evaluation metrics}

As for the acoustic units, they are evaluated in terms of their \textbf{bitrate}, where each unique embedding value is counted as a single symbol type. 
For the \textbf{bitrate} computation, each vector is processed as a character string. A dictionary of the possible values is constructed over the embedding file for the submitted test set. 
We thus assume that the entire test set corresponds to a sequence of vectors $U$ of length $n$: $U=[u_1,...,u_n]$. The bit rate for $U$ is then $B(U)=\frac{n}{D(U)} \sum_{i=1}^{n}{p(u_i)log_{2}p(u_i)}$, where $p(u_i)$ is the probability of symbol $u_i$ in $U$. The numerator is $n$ times the entropy of the symbols, which gives the optimal number of bits needed to transmit the sequence of symbols $s_{1:n}$. To obtain a bitrate, we divide by $D(U)$, the total duration of $U$ in seconds.\footnote{A fixed frame rate transcription may have a higher bitrate than a ``textual'' representation due to the repetition of symbols across frames. For instance, the bitrate of a 5 ms framewise gold phonetic transcription is around 450 bits/sec and that of a ``textual'' transcription around 60 bits/sec.} Not reported here, Task 3 also included a version of ABX in which the minimal triphones (``fly'' versus ``fry'') were extracted and presented as small audio clips, in order to not penalize the evaluation of sequence-to-sequence systems that would lack the alignemnt between units and audio signals. As it happended, no such system has been submitted as yet. 

As for the generated waveforms, native speakers of the test languages were recruited online to evaluate the quality of the synthesis in terms of intelligibility, naturalness, and similarity. \textit{Intelligibility} was measured by asking participants to orthographically transcribe the synthesized sentence. Each transcription was compared with the gold transcription using the Levenshtein distance, yielding a Character Error Rate (\textbf{CER}).  The overall \textit{naturalness} of the synthesis was assessed on a 1 to 5 scale, yielding a Mean Opinion Score (\textbf{MOS}).\footnote{The question posed was: \emph{Rate how natural the audio is, between  1 and 5 (1=very unnatural, 3 = neutral, 5=very natural).}} \textit{Speaker similarity} was assessed using a 1 to 5 scale. Sentences were presented in pairs (target voice, system voice).\footnote{The question posed was: \emph{Rate the similarity between the reference
 voice and the system voice, between 1 
and 5 (1 = very different voices,
 3 = neither similar nor different voices,
 5 = very similar voices).} Ten additional trials were included, for each participant, in which the reference voice was not the target voice but the source voice.} 
Each sentence token was evaluated at least once with each system,  as well as the original (gold) recordings.

\subsubsection{Datasets} The 2019 Benchmark for Task 3 (TTS0-19) provides training and test data for two language: English (the dev language) and Indonesian (the test language). For each language, one "Unit" dataset is provided to train unit discovery (around 15h, betwee, 100 and 120 speakers), and one "Voice" dataset is provided to train speech synthesis in the target voice (two voices in English, one in Indonesian, between 1h30 and 2h30 per speaker). The test set consists in new utterances by new speakers (around 30 minutes, 15-24 speakers; see Table \ref{tab:zrc_datasets} for detailed numbers). None of these datasets are provided with labels except an anonymous speaker ID.

\subsubsection{Baselines}


The baseline system consists of an existing  acoustic unit discovery system  which discovers GMM--HMM models and clusters them using an unsupervised Bayesian approach (see \cite{DBLP:conf/sltu/OndelBC16}). We then use decoding from this system (i.e.,  sequences of unit labels) instead of phonemes, and train an out-of-the-box speech synthesizer (Merlin, with the Ossian front end \cite{DBLP:conf/ssw/WuWK16}). For the topline system, we replace the  unit discovery with an off-the-shelf GMM-HMM ASR system.

\subsubsection{Results}

The performance was overall quite good, with several systems achieving better resynthesis than the text-based topline. As shown in Figure \ref{fig:MOSbitrate}, there is a general tradeoff between synthesis quality and bitrate, which held both in the dev language (English) and in the heldout surprise test language (Indonesian). As shown by the black point in the figure (the decoded output of a simple phone recognizer), phonemic transcription is a highly-compressed representation of speech which is excellent for this task (the middling MOS scores are, as for Task 1, attributable to the fact that the out-of-the-box ASR and TTS were not optimized to the task).

Many of the systems that have a \textbf{low bitrate} (under 100b/sec)  learn a discrete autoencoder on acoustic features (\textbf{Kam19a--b,Yus19,Gök19,Liu19a--b,Gün20}), generally taking  further steps such as filtering or downsampling to reduce the temporal resolution. Taking a slightly different approach, our baseline model, as well as the related \textbf{Yus20a\nobreakdash--c}, learn latent HMMs as acoustic units, in order to explicitly model duration. On the other hand, \textbf{Pan19a--b,Kum20a,b}  put  temporal reduction in an initial step of acoustic segmentation based on syllable-like units. Among these models, \textbf{Kum20b}, which presegments and then learns HMM acoustic units, stands out as reaching performance comparable to higher bitrate models (it admittedly has a somewhat higher bitrate than the other models listed here). Syllable-like presegmentation, as noted above, has also been used productively in Task 2 by \textbf{Ras17, Alg22}. It is fair to say that syllables have been underutilized in zero resource speech processing, given their promise.

Most of the remaining systems have a \textbf{high bitrate} between 100 and 600b/sec. Supervised posteriorgrams are on the upper end of this, and MFCC representations have a bitrate around 1500.\footnote{Note that our bitrates are calculated empirically on a rather small speech corpus. The representational \emph{capacity} of MFCC representations is an order of magnitude higher.} Most of the submitted systems in this range are compression approaches using discrete autoencoders, including the system of \textbf{Che20b}, which gives excellent performance. The system of \textbf{Nie20a,b} stands out among the others as yielding high quality results. This is the only submitted system which uses a predictive loss based on CPC---although, unlike typical CPC models, it works from spectrogram and is trained on the  small (15h) dataset provided for the 2019 edition.

The results of \cite{lakhotia2021generative} also support the claim that CPC and related approaches are well-adapted to discrete resynthesis.  In addition,  \cite{lakhotia2021generative} demonstrated that an automatic evaluation using ASR is strongly predictive of human evaluators' ratings, and that the discrete representations can be used to support learning a language model.

\subsection{Task 4: spoken LM}
\label{sec:t4}

Spoken language modeling is the task of learning a language model directly from  audio.  Such a model could be end-to-end, learning directly from speech, or it could take as input discrete or continuous representations from Task 1 or word level representations from Task 2---so long as these input representations are learned without supervision from text or other labels. The task can be understood as the modeling of the probability distribution of spoken utterances in an unknown language.

\subsubsection{Evaluation metrics}
For Task 4, the evaluation problem is severe. Language models trained from text are typically evaluated by the perplexity over a test corpus, or by fine-tuning on downstream tasks. As discussed above, the ZRC series focuses evaluation on zero-shot tasks that require no training. This excludes a fine-tuning evaluation. As for perplexity, in text-based systems, it is derived from the conditional probability distribution of the next token given a past sequence of tokens. In speech-based systems that use discrete pseudo-text units, the number of such units is a latent variable, making the perplexities difficult to compare across models. The problem becomes worse for systems that do not use discrete representations at all, where the estimation of the conditional probabilities themselves becomes model dependent. The two editions of the ZRC 2021 used a battery of 4 metrics, each one measuring performance at a different linguistic level: acoustic, lexical, syntactic and semantic. 

At the level of \textit{acoustics}, we the ABX-LS benchmark as defined in Section \ref{sec:t1}, built on top of LibriSpeech dev and test sets (see \cite{Kahn_2020_librilight,riviere2020unsupervised}). 

At the \textit{lexical} and \textit{syntactic} levels, instead of computing an average perplexity across a corpus, the ZRC uses a contrastive approach, where a ``pseudo-probability'' $\hat{p}$ is computed for minimal pairs of utterances---one grammatically legal, the other illegal. The pseudoprobability can be obtained from a language model by decomposing the probability of an utterance $U$ into a product of conditional probabilities of each of its constituant units $u_i$: $\hat{p}(U)=p(u_1)p(u{_2}|u{_1}) .. p(u_N|u_1..u_{n-1})$, or by computing an average perplexity score or of the loss function over the utterance $U$. An accuracy $Acc$ is computed by counting how often $\hat{p}$ is higher for the legal than for the illegal utterance: 
$$Acc(T)=\frac{1}{|T|}\sum_{(a,b) \in T}{(\mathds{1}_{\hat{p}(a)>\hat{p}(b)}+\frac{1}{2}\mathds{1}_{\hat{p}(a)=\hat{p}(b)})}$$

where $T$ is a test set containing pairs of audio, one legal ($a$), one illegal ($b$); the chance level is 0.5. 
To probe the lexical level, pairs of well matched words versus nonwords (e.g. ``brick'' versus ``blick'') are constructed using the Wuggy nonword generator \cite{keuleers2010wuggy}. The syntactic levels is probed by using pairs of grammatical and non grammatical sentence derived from the BLIMP dataset \cite{warstadt2019blimp}. All stimuli are synthetized using the Google TTS API resulting in the sWUGGY and sBLIMP test sets, respectively (see below for details).

Finally, ZRC evaluated the \textit{semantic} level by using a similarity {probe task} used to investigate word embeddings \cite{chung2018speech2vec}. It correlates the similarity of systems' representations of words with human similarity judgments. This enables us to measure the extent to which the model is able to extract lexical semantic knowledge. 
As for the ABX task, participants provide embeddings for input tokens as well as a distance to compute similarity. The Spearman rank correlation is calculated between the dissimilarity scores provided in the submission test set (sSIMI), $d(a,b)$, and the dissimilarity scores given by human judgments, $d_h(a,b)$. The challenge provided by default the cosine distance computed over pooled embeddings (with mean, max or min pooling).

\subsubsection{Datasets} The there is only one benchmark (sLM-21) associated with Task 4. The default training set is LibriSpeech 960h, although participants can use other training sets so long as no labels is provided besides speaker ID. The test set is split into sWUGGY, sBLIMP and sSIMI, that evaluate the lexical, syntactic and semantic levels, respectively. These test sets are described in details in \cite{nguyen2020zero} and only briefly summarized here. 

The sWUGGY dev and test sets consists of 5,000 and 20,000 pairs of words and nonwords respectively, with the existing words being part of the LibriSpeech train vocabulary. There is also an additional OOV-sWUGGY dev and test set sconsisting of 5,000, and 20,000 pairs respectively, with existing words which do not appear in the LibriSpeech training set. The nonwords are produced with WUGGY \cite{keuleers2010wuggy}, which generates, for a given word, a list of candidate nonwords best matched in phonotactics and syllabic structure, which were additionally filtered for pronouncability using G2P, and for having on average the same unigram and bigram phoneme frequencies as words. Waveforms were produced with the Google Speech API. 

The sBLIMP dev and test sets are adapted from BLIMP \cite{warstadt2019blimp}, a set of linguistic minimal sentence pairs of matched grammatical and ungrammatical sentences. The dev and test sets contain 6,300 and 63,000 pairs respectively, with no sentence pair overlap. Stimuli were filtered to contain LibriSpeech vocabulary and for natural prosodic contours, and synthesised as above.

The sSIMI dataset was constructed out of 13 existing semantic similarity and relatedness datasets: WordSim-353 \cite{yang2006verb},
WordSim-353-SIM \cite{agirre2009study}, mc-30 \cite{miller1991contextual}, rg-65 \cite{rubenstein1965contextual}, Rare-Word (or rw) \cite{luong2013better}, simLex999 \cite{hill2015simlex},
simverb-3500 \cite{gerz2016simverb}, verb-143 \cite{baker2014unsupervised} , YP-130 \cite{yang2006verb}
and the relatedness-based datasets include MEN \cite{bruni2012distributional}, Wordsim-353-REL \cite{agirre2009study}, mturk-287 \cite{radinsky2011word}, and mturk-771 \cite{halawi2012large}.
All scores were normalised on a 0-10 scale, and pairs within a same dataset containing the same words in different order were averaged.  Pairs containing a word absent from LibriSpeech train set \cite{Panayotov2015librispeech} were discarded. The mturk-771 dataset was set aside as a dev set and the other 12 datasets were used as test sets, after removing overlapping pairs across dev and test sets. Given the unequal size of the test sets, the ZRC Benchmark introduced a weighted average of the Spearman scores which we report here.
Two subsets of audio files, one synthetic, and one natural, were created, the latter being obtained by extracting the audio sequences corresponding to each word from LibriSpeech, as in \cite{chung2018speech2vec}. In this subset, each word can appear in multiple tokens, providing phonetic diversity; duplicated \textit{scores} are averaged in the analysis step. The natural subset is smaller than its synthetic counterpart, as we had to discard pairs from the test and dev sets which were not present in the LibriSpeech test and dev sets respectively.  The synthesized subset is composed of 9744 and 705 word pairs for the test and dev sets respectively, and the LibriSpeech subset is composed of 3753 and 309 pairs for the test and dev sets.

\subsubsection{Baselines}
Our baseline system, described in \cite{nguyen2020zero}, is based on first training a contrastive predictive coding (CPC) model. 
We review the CPC acoustic model \cite{oord2018cpc} here for clarity.
Given an input waveform $\textbf{x}$, the \emph{encoder} component of the model maps it to a sequence  $\textbf{z}=(z_1,\dots,z_T)$. An autoregressive component then predicts the future, taking  $z_1,\dots,z_t$ and outputting a latent representation $c_t$, which is a representation of the context. Given the context $c_t$, CPC tries to predict the $K$ next future embeddings $\{z_{t+k}\}_{1\leq k\leq K}$ by minimizing a constrastive loss:
\begin{equation}\label{eq:CPCloss}
    \mathcal{L}_t = - \frac{1}{K}\sum_{k=1}^K\log \left[\frac{\exp \left(z_{t+k}^\top W_kc_t\right)}{\sum_{\Tilde{z}\in\mathcal{N}_t}\exp \left(\Tilde{z}^\top W_kc_t\right)}\right]
\end{equation}
where $\mathcal{N}_t$ is a random subset of negative embedding samples, and $W_k$ is a linear classifier used to predict step $k$ of the future. 

Our baseline system then clusters the resulting framewise representations (as independent observations) using k-means, to reduce them to 50 units. The resulting discrete sequences are passed as input to a character-based language model. We experimented with both BERT and LSTM models, and found that large BERT models performed best.

\subsubsection{Results}

The first round of submissions was documented in 2021 \cite{dunbar2021zero}; the best-performing systems were variants of our baseline system.
A second round was opened as a NeurIPS 2021 challenge, including a visually-grounded training option. 
Briefly, this modified scenario expands the range of data that models can be trained on, to include multi-modal datasets (like speech and image, or speech and video). The rationale is that young children learn in a multimodal, multisensory enviroment rather than by just listening. Some earlier models of word discovery and representation learning  demonstrated the feasibility of such muldimodal training  \cite{harwath2016unsupervised,arandjelovic2017look,chrupala2017representations}. Following \cite{alishahi2021zr}, Task 4 was expanded to  include ``visually-grounded'' training. 
Participants were to indicate the dataset they used. Systems were only tested with speech-only inputs, however, for comparability with non grounded systems.
Here, we present for the first time the results of these latest submissions to Task 4 (see Table \ref{tab:res2021b}). 

\newcommand{\gr}{\rowcolor[HTML]{E0E0E0}}
\begin{table*}[ht]
\caption{\textbf{Leaderboard update for December 2021.} Test set results on the four metrics for acoustic (ABX), lexical (sWUGGY), syntactic (sBLIMP) and semantic (sSIMI) levels respectively. The first three systems are for reference  and include the two best systems of the first 2021 ZRC edition \cite{dunbar2021zero}. The last three systems are \textbf{visually grounded (VG)}. The remaining systems only use audio input. The best overall score in each column among all submitted systems is bolded.\label{tab:res2021b}}
\centering
\begin{tabular}{cccccccccc} 
&    &  \multicolumn{2}{c}{ABX-with.} &  \multicolumn{2}{c}{ABX-across}& sWUGGY & sBLIMP  & \multicolumn{2}{c}{sSIMI}   \\
\cline{3-4}\cline{5-6}\cline{9-10}
System  & Budget & clean  & other & clean  & other  & & & synth.   & Libri. \\ \hline\hline
Character-based (RoBERTa) & 24576       & -      & -     & -      & -     & {96.25}   & {82.11}  & {23.22}  & {21.08}  \\
\gr Best March 2021 Low-budget \textbf{[Nie21]} & 60 &5.41  &{8.67}&{6.89}&{13.14}&{72.86} &{53.59}  &{5.60}&2.72 \\
    Best March 2021 High-budget \textbf{[BAS4-lg]}  & 1536 &3.28 &4.81&{4.31}&{7.92}&{75.51} &{56.16}  &3.19&1.32\\ \hline\hline

    Gan21  & 60  & 5.04 & 7.89 & 7.08 & 14.01 & 59.87 & 52.20 & 7.47 & 4.38\\
\gr Ngu21a & 2560  & 3.49 & 5.72 & 4.56 & 9.19 & 77.22 & 55.62 & 8.43 & 4.48\\
    Ngu21b & 2112  & 3.80 & 4.86 & 4.70 & 7.70 & 79.13 & 58.89 & 8.62 & 6.65\\
\gr Ngu21c & 3424  & {3.03} & \textbf{3.62} & 3.83 & \textbf{5.63} & 80.19 & 59.29 & 6.8 & 3.08\\
    Ngu21d & 2720  & 3.49 & 5.72 & 4.55 & 9.20 & \textbf{80.29} & \textbf{59.93} & 9.43 & 5.33\\
\gr Lan21  & 60  & \textbf{2.85} & 4.44 & \textbf{3.67} & 7.33 & 70.23 & 51.93 & 3.09 & 2.05\\
    Bha21a & 60  & 3.49 & 5.72 & 4.51 & 9.16 & 50.90 & 52.10 & 3.44 & 1.67\\
\gr Bha21b & 60  & 3.49 & 5.73 & 4.57 & 9.19 & 48.55 & 52.80 & 1.57 & 5.2\\
    Gao21a & 60  & 6.71 & 10.62 & 8.41 & 15.16 & 62.29 & 53.17 & 1.67 & 1.13\\
\gr Gao21b & 60  & 6.60 & 10.86 & 8.51 & 16.51 & 68.42 & 54.02 & 2.61 & 1.78\\
    Gao21c & 35  & 7.45 & 11.76 & 10.94 & 19.31 & 62.29 & 53.17 & 5.07 & 1.2\\
\hline
\gr Pen21  (VG)  & 468  & 4.24 & 5.22 & 5.08 & 7.91 & 75.23 & 57.40 & \textbf{17.99} & \textbf{12.78}\\
    Lee21a (VG) & 7  & 6.50 & 9.95 & 9.17 & 15.46 & 51.91 & 50.43 & 4.51 & 5.54\\
\gr Lee21b (VG) & 7  & 6.50 & 9.95 & 9.04 & 15.44 & 51.91 & 50.43 & 8.1 & 12.21\\ \hline
\end{tabular}
\end{table*}

Similar to the baseline models, the systems of \textbf{Gan21,} \textbf{Ngu21a,d} \textbf{Bha21a,b,} and \textbf{Gao21a\nobreakdash-c} take the approach of training acoustic units and then constructing a language model on their outputs. The distinction between \textit{high-budget} systems and \textit{low-budget} systems is made the basis of the number of GPU hours needed to train the language model.  \textbf{Gao21a\nobreakdash-c} apply segmentation and pooling to reduce the temporal resolution of the units, while \textbf{Bha21a,b} use Segmental CPC to learn units and segmentation jointly. \textbf{Ngu21a,d} are technical improvements on the previous best system \textbf{BAS4-lg}. On the other hand, \textbf{Ngu21b,c} are HuBERT systems, trained end-to-end on a masked language modelling task.

The systems of \textbf{Pen21} and \textbf{Lee21a,b} are visually grounded. In the case of these two systems, that means they both start from acoustic units that are trained using parallel speech--image data (picture captions). One  difference between the two models is the type of training---\textbf{Pen21} trains end-to-end on a masked language modelling objective, while \textbf{Lee21a,b} use the pre-trained features as input to a small BERT.

Task 4 is clearly in its very early stages (this in spite of the excellent ABX performance of the units used in systems up to now). However, even at this stage, after only one year's worth of submissions, spoken language modelling has shown improvement on the spot-the-word task (moving from the best speech-based baseline's 75\% accuracy up to 80\%) and on the syntactic judgment task (improving from 56\%  to 60\% accuracy). The approach so far has been simple: high-quality units and a powerful language model. In the baseline models as well as most submissions, these components were trained separately; newer models like HuBERT \cite{hsu2021hubert}  learn them jointly. The two approaches are currently tied for the top position on the leaderboard (\textbf{Ngu21a,d}, CPC units fed to a large BERT model, and \textbf{Ngu21b,c}, HuBERT systems). As for capturing word semantics, the Fast-VGS+ system of \textbf{Pen21} stands out as a serious competitor. This visually-grounded system takes advantage of spoken image caption data in training. 


\section{What next?}
\vspace{-0.1em}
Over the  six editions of the ZRC series, the following lessons can be drawn:
\begin{itemize}
\item Great progress has been made in Acoustic Unit Discovery (T1) due to recent breakthroughs in self-supervised representation learning  showing good scaling properties in large corpora. The latent units discovered at this stage, though, are not interpretable linguistic units like phonemes but represent shorter-duration acoustic events. 
\item The Discrete Resynthesis task (T3) obtained excellent results, sometimes surpassing text-based systems in resynthesis quality---at the expense of bitrate, which remains about 4 to 8 times higher than phonemes or text.
\item Spoken Word Discovery (T2) still remains disappointingly difficult in all of its three subcomponents (matching, clustering, and segmentation). Presumably, understudied effects of the acoustic and temporal variability inherent to speech still hampers current approaches. \item Spoken Language Modeling (T4) got  surprisingly promising results, given that the task is complex, considering the difficulties found in Task 2, and the fact that most systems only worked from Task 1 units with sub-phonemic temporal granularity. There is however room for progress, given the gap between speech-based and text-based language models on syntactic and semantic tests.
\end{itemize}


\label{sec:questions}
Given the large body of results that have accumulated, it may now be useful to reflect on some of the basic assumptions and methods of the ZRC series to determine how to move forward. The assumptions are related to the architecture presented in Figure \ref{fig:blueprint}b and its corresponding  task decomposition. The methods are related to the particular choice of metrics that were chosen to evaluate each of these tasks. We discuss in particular the role of Acoustic Modeling and the Lexicon.  

\vspace{-0.9em}
\subsection{Acoustic Modeling and ABX.}
\vspace{-0.1em}

One of the basic assumptions of the ZRC series is the existence of an Acoustic Modeling component that turns speech input into a latent representation, which plays the role of phonemes or text in that it can be directly used as input to other processing components: the Lexicon, Waveform Generation, and possibly even Language Modeling. Methodologically, we proposed the machine ABX task as a metric to gauge the quality of this latent representation. This makes the prediction that there should be a strong positive correlation between ABX scores and the relevant metrics in the other components. Inspection of this correlation across tasks shows that such a correlation exists, but that it is in some cases weak. 

\begin{itemize}
    \item 
\textbf{T1 and T3}: a reanalysis of the 35 ZRC submissions shows a Pearson correlation coefficient of $r=.57$ and $r=.54$ (English and Indonesian, resp.) between ABX and intelligibility as measured by Character Error Rate (CER). The systems differed not only in the encoder but also in the decoder introducing noise in the correlation. A more controlled correlation across 9 systems with matched decoders \cite{lakhotia2021generative} reported $r=.905$.\footnote{There is also a weaker (negative) correlation of ABX and bitrate (-.156 and -.277, in English and Indonesian, resp.) suggesting that a resynthesis objective may push away from a low bitrate phoneme-like representation.} 

\item \textbf{T1 and T4}: we also find a reasonably high correlation between ABX and spot-the word ($r=.52$ across the ZRC submitted systems, and $r=.853$ in \cite{lakhotia2021generative} in a more controlled comparison with matched language models). 

\item \textbf{T1 and T2}: the situation is confusing. Since the beginning, we have seen that these two tasks may require different representations. For instance, unit discovery worked well with MFCC, but word discovery worked better with PLP. Similarly, \cite{algayres2020evaluating} showed that, across 16 types of word embeddings (supervised or unsupervised), ABX scores correlate only moderately well with two other proxies for word segmentation (frequency estimation: .53; Mean Average Precision: .45). 
\end{itemize}

While the observed level of correlation may be sufficient to still use ABX as a proxy for comparing models before using them for downstream tasks, some caution is necessary, especially for the link between T1 and T2. One possible explanation for this discrepancy is that the assumption of a single acoustic level feeding all downstream tasks is wrong, and that there are instead several different acoustic codes with different properties. Alternatively, it could be that there is a single code, but that the ABX metric is not capturing the linguistic proprerties of this representation relevant for all the other tasks. While ABX was constructed to measure contrasts in minimal pairs of possible word across changes in speaker, it may not capture well other kinds of invariance (speaking rate, phonetic context) that are crucial for some of the other tasks.  Further studies will be needed to sort out this question.

\vspace{-0.9em}

\subsection{The Lexicon, discrete units and interpretability}
\vspace{-0.3em}


One reason text-like representations---be they phonemic, alphabetic, or logographic---are fundamental to speech and language processing is that they serve a dual function. On the one hand, they record linguistically important properties of the \textit{form} (what was actually uttered). On the other hand, they support straightforward analysis of the \textit{content} (``meaningful'' properties like morphology, syntax, and semantics).

While human listeners are sensitive to detailed, sub-phonemic properties \cite{gwilliams2018spoken}, and while various gradations in lexical meaning can be observed \cite{beekhuizen2021probing}, the two kinds of variability are not generally correlated. For example, although it is possible to pronounce the noun \emph{sun} with an initial sound that would be intermediate between an /s/ and a /f/---making it sound somewhat more like the adjective \emph{fun}---this gradient change does not evoke a concept of ``slightly amusing star,'' nor make the word more adjective-like. In other words, text-like representations would thus seem to be necessary for decorrelating form from meaning using an arbitrary mapping between a word's phonological forms and its semantic or syntactic representations \cite{desaussure1916course}. 

However, achieving this crucial decorrelation may not necessarily require that the representations be discrete, nor that they correspond to interpretable linguistic units or ``words'`' as defined in a dictionary. First, linguistically, there are at least three distinct notions of `words' (prosodic, syntactic and semantic  \cite{van2014word}), which may not may not be aligned with how dictionaries are constructed.\footnote{As an example, ``a piglet'' could be considered a single unit prosodically, two units semantically, and three units, morphosyntactically (``a+pig+let'').} Second, dictionaries are the result of a long cultural evolution where many design choices have been made that may not be consistent within or across languages. As a result, it could be that the requirement for T2 to provide segmentations and lexicons that are aligned with the written text is too strong. The fact that word-based units like BPE work well for text-based applications does not mean that the equivalent units for speech would align well with word boundaries. We could imagine in the future replacing the T2 metrics by new metrics reflecting the functional role of word segmentation, i.e., that of providing a level of granularity where arbitrary mappings between form an meaning can be learned (along the lines of the sSIMI metric in T4).

\subsection{The future of the Zero Resource Speech  Challenge}\label{sec:openchallenge}

The submission site, \url{www.zerospeech.com}, is now open continuously and allows for 
running evaluations on all of the past benchmarks. 
The field of unsupervised representation learning is 
established enough that it is no longer necessary to channel it through special events. Indeed, self-supervised audio models are such an active domain that there are many relevant new models (for example, WavLM: \cite{chen2022wavlm}) which have yet to be evaluated on the ZRC metrics. Existing benchmarks, especially for Tasks 2 and  4, also still have a lot of potential for improvement, without creating more difficult tasks. 

One exception to this is shown in Figure \ref{fig:blueprint}. Combining Task 3 and Task 4 leads  naturally to consider the possibility of \emph{generating} spoken language. A traditional spoken dialogue system will (conditional on some knowledge source) generate text, typically using a neural language model. The text is  synthesized into speech. A spoken language model can be made to generate spoken language directly, as demonstrated by \cite{lakhotia2021generative,kharitonov2021pgslm,nguyen2022dialoggen,qian2022contentvec}. Much as Task 3 is complementary to Task 1---but has slightly different constraints---the task of generating speech from a spoken language model is complementary to Task 4, yielding a  potential Task 5. The evaluation of such a potential task may follow \cite{lakhotia2021generative}, replacing human evaluations of  intelligibility and meaningfulness of by ASR-based proxy measurements (Phone Error Rate for intelligibility in a Task 3 setting, and continuation BLEU or VERT score for the prompted or unmprompted generations). 

Another reason for not declaring the ZRC series closed is that there is still a lot to understand on the evaluation side for existing tasks. For Task 1, recent research has shown that many discovered representations may not be speaker invariant \cite{de2022probing} and  differ from how humans perceive speech sounds \cite{millet2020perceptimatic,millet2021predicting}. Further, their ability to perform out of domain (noisy enviroments, accented speech) has not been evaluated \cite{kharitonov2021data,riviere2021towards}. 

\section{Conclusion: Towards textless NLP}
 
Research on Acoustic Unit Discovery has led to a wave of new models using unsupervised pre-training to advance ASR. It also opens up the more radical possibility that one may get rid of text altogether, and proceed with building language processing pipelines directly from raw audio. Up to now, this possibility has done little to change the dominance of text as the basic currency of NLP. One  exception is translation, in which the idea of  training machine translation directly from speech to speech in an end-to-end fashion has seen substantial uptake \cite{tjandra2019speech,jia2019direct,lee2021direct,jia2021translatotron,lee2021textless}. Other ways of removing text from the processing stack have also been explored, such as for language generation \cite{lakhotia2021generative}. In general, however, converting speech to text and back remains the first and the last step of  current speech-based NLP systems.  

The allure of replacing text with low-bitrate unsupervised representations of speech goes beyond bringing NLP to lower-resource languages. Learned pseudo-text promises to be more flexible than traditional orthography: if the transcription system is the result of learning, it can also change to deal with new varieties and accents, and can learn to capture important linguistic information which is captured in a very limited way by typical writing systems, such as prosody. On the other hand, it promises to be more consistent:  some writing systems are complicated by arbitrary exceptions, while other languages lack standardized conventions for spelling. Discovered representations could avoid both of these issues. Unlike traditional phonetic transcription, which uses a fixed, universal set of symbols which can in fact have rather different phonetic values across languages, unit discovery allows for a system to be adapted to the language.

The Zero Resource Speech Challenge has spearheaded efforts to build demonstrably useful unit discovery, as well as stimulating  progress  in applying these representations to more complex tasks. The major advances in building more realistic auditory-like representations have already borne fruit in recognition and synthesis. As we move toward better evaluations, we look ahead to the possibility of truly textless NLP---and a major key to unlocking cognitive models of human language development and speech perception.

\section*{Acknowledgment}

For E. Dunbar, this research was supported by the Connaught Fund and  the Arts and Science Bridging Fund, U. of Toronto, Natural Sciences and Engineering Research Council of Canada (NSERC) RGPIN-2022-04431, and by ANR grant ANR-17-CE28-0009 GEOMPHON. For E. Dupoux in his EHESS role and N. Hamilakis, it was supported by ANR grants ANR-17-EURE-0017 Frontcog, ANR-10-IDEX-0001-02 PSL* and ANR-19-P3IA-0001 PRAIRIE 3IA Institute, and by a Meta AI Research Gift.

\bibliographystyle{IEEEtran}
\bibliography{zr}

\begin{IEEEbiography}[{\includegraphics
[width=1in,height=1.25in,clip,
keepaspectratio]{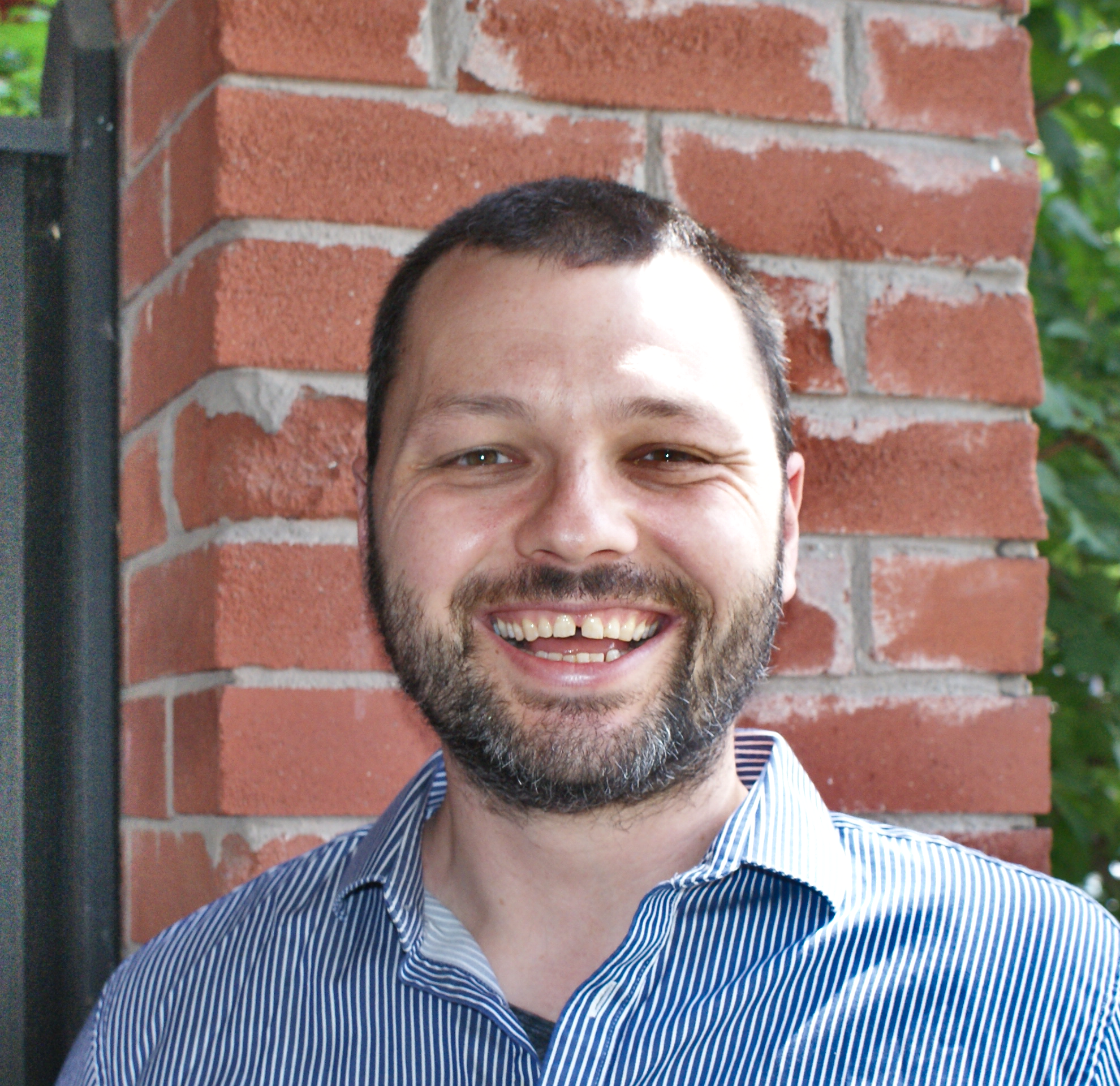}}]
{Ewan Dunbar}
is an Assistant Professor in the French Linguistics program at the University of Toronto and an Affiliated Scientist with the Cognitive Machine Learning (CoML) team at the Ecole Normale Supérieure (ENS) in Paris and INRIA. He holds a PhD in Linguistics from the University of Maryland, College Park, an MA in Linguistics from the University of Toronto, and a BSc in Linguistics and Computer Science, also from the University of Toronto.
His research focuses on speech perception, psycholinguistics, and speech technology.   He is co-organizer of the Zero Resource Speech Challenge (2017--2021).
\end{IEEEbiography}

\begin{IEEEbiography}[{\includegraphics
[width=1in,height=1.25in,clip,
keepaspectratio]{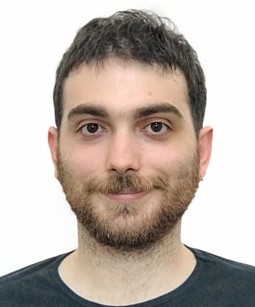}}]
{Nicolas Hamilakis}
is a Research Engineer for the Cognitive Machine Learning (CoML) team at the Ecole Normale Supérieure (ENS) in Paris and INRIA.  He obtained an undergraduate degree  in mathematics and computer science at the Pierre \& Marie Curie University and an Masters in the engineering of programs and algorithms at Université de Paris.
His work is a mix of data management and software engineering of experimental research tools. His interests are in software engineering and in  how to provide better, more elegant solutions and reproducible results. He worked on the organizational committee of the 2021 edition of the challenge including the construction of a pipeline for evaluating  submissions.  
\end{IEEEbiography}

\begin{IEEEbiography}[{\includegraphics
[width=1in,height=1.25in,clip,
keepaspectratio]{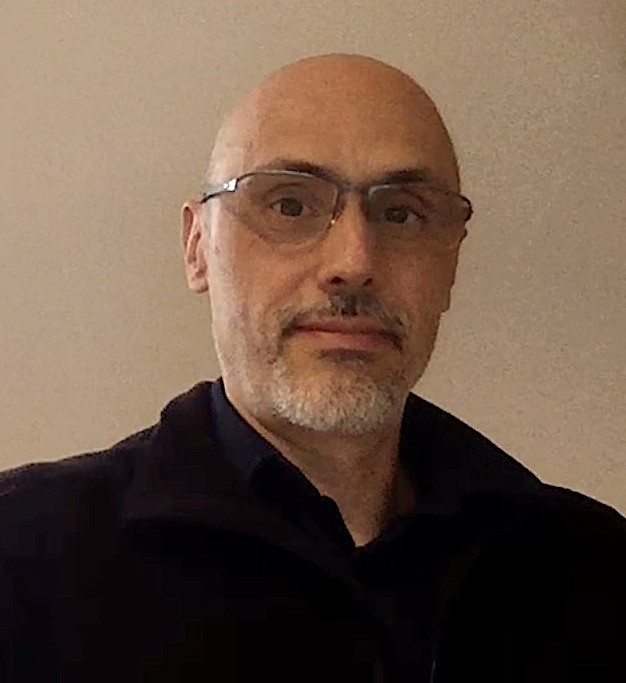}}]
{Emmanuel Dupoux}
is a Professor at the École des Hautes Études en Sciences Sociales (EHESS) and Research Scientist at Meta AI Labs. He directs the Cognitive Machine Learning (CoML) team at the Ecole Normale Supérieure (ENS) in Paris and INRIA.   His education includes a doctorate in Cognitive Science (EHESS), a Masters in Computer Science (Orsay University) and an undergraduate degree in Applied Mathematics (Pierre \& Marie Curie University).
He is 
co-organizer of the Zero Resource Speech Challenge series (2015--2021). 
He is a CIFAR LMB and a ELLIS Fellow.
 He has authored 150 articles in peer reviewed outlets in cognitive science and language technology. 
\end{IEEEbiography}

\clearpage
\pagebreak


\setcounter{section}{0}
\setcounter{page}{1}
\setcounter{figure}{0}
\setcounter{table}{0}
\renewcommand{\thesubsection}{S\arabic{subsection}}
\def\thesubsectiondis{S\arabic{subsection}.}
\renewcommand{\thesection}{S\Roman{section}}
\renewcommand{\thefigure}{S\arabic{figure}}
\renewcommand{\thetable}{S\arabic{table}}
\renewcommand{\thepage}{\roman{page}}

\end{document}


\title{Self-supervised language learning from raw audio:\\ 
\fontsize{22}{26}\selectfont lessons from the Zero Resource Speech Challenge series}

\author{
Ewan Dunbar, Nicolas Hamilakis, 
and Emmanuel Dupoux
\thanks{Ewan Dunbar is with the Department of French,
University of Toronto, Toronto, ON, M5S 1A1, Canada (e-mail:
ewan.dunbar@utoronto.ca).

Nicolas Hamilakis is with the École Normale Supérieure, 75005 Paris, France, and with Inria, 75013 Paris, France (e-mail: nick.hamilakis562@gmail.com).

Emmanuel Dupoux is with the École des Hautes Études en Sciences Sociales, 75006 Paris, France, and with Meta AI, 75009 Paris, France (e-mail: emmanuel.dupoux@gmail.com).}}


\markboth{Submitted to Journal of Selected Topics in Signal Processing}%
{Lessons from the Zero Resource Speech Challenge series}


\maketitle

\begin{abstract}
The Zero Resource Challenge series provides a blueprint for building a full dialogue system in a self-supervised or unsupervised way from raw audio without using any textual representations or expert labels such as phonemes, dictionaries or parse trees. The blueprint consists of breaking down this long-term objective in terms of well-defined tasks and metrics enabling model comparison and cumulative progress. So far, four tasks have been defined: Acoustic Unit Discovery, Spoken Term Discovery, Discrete Resynthesis, and Spoken Language Modeling. We present an overview of the six editions of this challenge series since 2015, discuss the lessons learned, and outline the areas which need more work or give puzzling results. 
We also discuss the future of the series, introducing a new task (generative spoken language modeling) and a new rolling challenge submission format. 
\end{abstract}

\begin{IEEEkeywords}
textless speech processing, unsupervised and self-supervised learning, representation learning.
\end{IEEEkeywords}

\section{Introduction}


\begin{figure*}[b]
\centering
\includegraphics[width=0.7\textwidth]{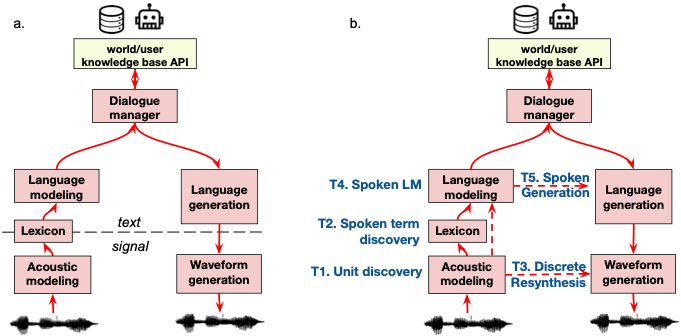}
\caption{a. Traditional pipeline for a spoken assistant based on textual resources b. Pipeline and tasks for the zero resource challenge series.}
\label{fig:blueprint}
\end{figure*}

\IEEEPARstart{F}{or} a long time, language technology has been developed principally using large quantities of textual resources. This makes sense, since, as far as technological applications are concerned, language has primarily been used in written form. When it comes to dealing with the spoken  language, however, this has given rise to a division of labor between, on the one hand, speech components which aim at converting speech to text or text to speech (ASR, automatic speech recognition, and TTS, text-to-speech synthesis), and, on the other hand,  components that perform a variety of language tasks based on text (language understanding, dialogue, language generation). As a result, even fundamentally speech-first applications like speech-to-speech translation or speech assistants like Alexa or Siri are cobbled together in a Frankensteinian fashion, with some components trained on text and others trained on speech (see Figure \ref{fig:blueprint}a)---and with all the speech components trained using large amounts of supervision (textual transcription) so that they can communicate with the text-based components. But is this a necessity? Could we build spoken-language based applications directly from the audio stream without using any text?

\begin{table*}[]
\centering
\caption{\textbf{Summary of the metrics and tasks used in the Zero Resource Challenge series}. $d$ is a dissimilarity measure between embeddings ($d_h$ is from human judgments). $\hat{p}$ is a pseudo-probability computed by the LM over the entire input sequence. $ED$ is the edit distance over the phonetic transcriptions of the dicovered segments. $D(U)$ is the duration of utterance $U$ (in sec).}

\label{tab:tasks}
\def\arraystretch{1.2} 
\begin{tabular}{p{3.2cm}p{1.5cm}p{4.0cm}p{3.5cm}p{3.7cm}}
\hline
\rowcolor[HTML]{EEEEEE} \bf Task  & \bf Ling. Level             & \bf Metric & \bf Model outputs examined &\bf Example   \\
\hline
T1. Unit Discovery & Phonetic & 
\textbf{ABX}: $d(a,x)<d(b,x)?\newline a\in A, b\in B, x\neq a\in A $ & triplets of frame embeddings & /bit$_{T1}$/,  /bet$_{T1}$/, /bit$_{T2}$/ \\ \hline
  & Lexical (matching) & \textbf{NED} : $ED(a,b)/max(|a|,|b|)$ \newline \textbf{COV}: fraction of corpus covered & pairs of speech segments\newline (segment:pair of time-stamps) & a $[$\textbf{rose}$]$ is a $[$\textbf{rose}$]$ ...\\  \cline{2-5}
T2. Spoken Term Discovery & Lexical (clustering) & \textbf{Grouping F-score} \newline \textbf{Type F-score}  & clusters of speech segments & a $[$\textbf{rose}$]$ is a $[$\textbf{rose}$]$ is a $[$\textbf{rose}$]$\\ \cline{2-5} 
    & Lexical\newline(segmentation) & \textbf{Token F-score}: as defined in text  \newline \textbf{Boundary F-score}: segmentation  & {list of time-stamps} & a$|$rose$|$is$|$a$|$rose$|$isa$|$ro$|$se$|$ \\ \hline 

T3. Unsup. Discrete Resynthesis ("TTS without T") & Phonetic & \textbf{Bitrate}: $\frac{n}{D(U)}\sum{p(s_i)log( p(s_i)) }$ \newline \textbf{MOS}: human evaluation& series of discrete units\newline waveforms & $U=s_{1},..,s_{n}$\\ \hline
   & Lexical\newline(frequency) & \textbf{spot-the-word}:  $\hat{p}(a)\textgreater{}\hat{p}(b)?$ & pairs of (pseudo) probabilities & $\hat{p}$(brick), $\hat{p}$($^*$blick)\\  \cline{2-5}
T4. Spoken LM   & Lexical\newline(semantics) & \textbf{similarity}:  $d(a,b) \propto d_{h}(a,b) ?$ & pairs of word embeddings &  $d_{h}$(abduct, kidnap) : 8.63\newline $d_{h}$(abduct, rotate): 0.5\\  \cline{2-5}
& Syntax & \textbf{accept. judgment}:  $\hat{p}(a)\textgreater{}\hat{p}(b)?$ & pairs of (pseudo) probabilities & $\hat{p}$(dogs eat meat), $\hat{p}$($^*$dogs eats meat)\\ \hline
\end{tabular}

\end{table*}

Preschoolers around the world demonstrate clearly that it is possible to learn to naturally interact in language well before learning to read and write \cite{dupoux2018cognitive,bavin_cambridge_2009}. Written language is, rather, a tool for transcribing spoken or signed language. Indeed, many languages have no writing system at all, and many other language communities do not use the written form of their language much (reportedly, more than half the world's languages do not have a stable or widely used writing system).

The Zero Resource Speech Challenge focuses on spoken, rather than signed, languages. Reverse engineering the feat of learning a language from spoken input only is a fascinating scientific question. For spoken language technology, advancing this question would unlock a number of novel applications. For one thing, it would allow for applications that address the needs of languages that are entirely or mostly unwritten. Even in languages with large amounts of textual resources, learning language representations from raw audio would help capture the dimensions of language that are typically poorly represented in text (prosody, emotional and non-verbal vocalizations, and so on). Beyond helping to model these aspects of language, capturing unwritten oral expression could improve the ability of machine learning systems to deal with spontaneous speech, thereby unlocking the rich syntax and vocabulary of oral registers, which very often differ greatly from the written form. This would foster richer and more naturalistic human/machine interactions. 

Building a text-free dialogue system using only raw speech is a difficult machine learning challenge. It requires us to rethink the spoken language processing stack from the ground up. The Zero Resource Speech Challenge (ZRC) series has been designed to address two interlocking research problems: the task problem and the evaluation problem. 

The \textit{task problem} 
is to break down the ill-defined objective of ``learning to process spoken language  without  text'' into a series of well-defined sub-problems. The ZRC series follows the general architecture of a spoken digital assistant to define the learning problem implied by the training of each component---the acoustic model, a lexicon, language models, waveform generation, and so on. But instead of using phonemes, characters or words as an intermediate representation, the components are allowed to develop their own latent representations. For instance, instead of outputting phonemes or characters, the acoustic model is assumed to output \emph{acoustic units} which may or may not be discrete. Such an architecture (see Figure \ref{fig:blueprint}b) naturally gives rise to the following five tasks: (T1) Acoustic Unit Discovery, (T2) Spoken Term Discovery (T3), Unsupervised Discrete Resynthesis (T4), Spoken Language Modeling, and (5) Spoken Language Generation. These are the textless counterparts of well-known tasks: phone recognition (T1);  ASR (T2), TTS (T3), Language Modeling (T4), Text Generation (T5). We will review these tasks in turn in the following sections. 

The \textit{evaluation problem} is to define metrics that enable model comparison and cumulative progress for tasks that are defined only through a self-supervised  objective. For instance, ASR systems can readily be measured through phone or word error rates. But their self-supervised counterparts, Acoustic Unit Discovery systems, do not aim at recovering phonemes, but a latent representation. How can we evaluate theses systems?  Interest in some of the above mentionned tasks predates the ZRC series (see for instance \cite{kohonen_1988,varadarajan2008,huijbregts2011,lee2012,jansen_2013,badino2014}, for Task 1), but since each of the published papers used its own metrics, it was difficult (and still is) to compare systems and measure progress. The general strategy of the ZRC series to address this challenge is to develop zero-shot probe tasks that are inspired by human psycholinguistics, and which require no model retraining. The reasoning is that, since we want to probe for latent representations at various levels of a self-supervised system, it is best to not train any classifier on top of it. Otherwise, it would be unclear whether the performance obtained reflects the system under observation or simply the classifier. For each task, zero-shot metrics were developed that probe for the different levels of linguistic knowledge that the self-supervised system is supposed to have learned. They only require the extraction of information readily available in the system (for example, embeddings, pseudo-probabilities), and are computed by a separate fixed module which is identical across systems. The evaluation metrics that go with the tasks are listed in Table \ref{tab:tasks} and will be presented in more detail in the following sections.

Our contributions are as follows:
\begin{enumerate}
    \item We provide a comprehensive overview of the results obtained across the different tasks and metrics of the ZRC series since 2015.
    \item We discuss the lesson learned  and outline the areas that need more work or give puzzling results
    \item We discuss the future of the series, in particular regarding the task problem and the evaluation problem
\end{enumerate}

\begin{table}[]
\setlength{\tabcolsep}{3.9pt}
\centering
\caption{\textbf{Summary of datasets in the Zero Resource Challenge series.} LS960 and LS100 refers to librispeech 960h and 100h, respectively} \label{tab:series}
\begin{tabular}{lp{0.9cm}p{3.5cm}p{1.9cm}}
\hline
Chall.  & Tasks             & Train Data   & Test Benchmark\\
\hline
2015\cite{versteegh_2016}     & T1, T2   & English (Buckeye 5h),\newline Xitstonga (2h30) &ABX-15,\newline TDE-15  \\
2017\cite{dunbar2017zero}     & T1, T2    & English (45h), French (24h), Mandarin (2h30), German (25h), Wolof (10h) &ABX-17,\newline TDE-17   \\
2019\cite{dunbar2019zero}     & T3.  &  English (15h+4h40), Indonesian (15h+1h30) & TTS0-19\\ 
2020\cite{dunbar2020}     &  T1,T2,T3  &  reboot of ZR17, ZR19 & \\
2021a\cite{dunbar2021zero}    & T1,T4         &   English (LS960 or LS100)   & sLM-21 (ABX-LS, sWUGGY, \\
2021b    & T1,T4         &   idem plus speech coco      &sBLIMP, sSIMI)\\
\hline
\end{tabular}
\end{table}

\section{Past and present}

Six editions of the Zero Resource Challenge have been proposed over the years and are summarized in Table \ref{tab:series}. Each edition has explored a different combination of tasks and introduced different datasets. Overall, the six editoins have received a total of 115 submissions from 29 teams. In addition, several papers have been published using some of the Zero Resource benchmark metrics, which we also include in our review. 


\subsection{Task 1: Acoustic Unit Discovery}


The goal of acoustic unit discovery is to learn representations (embeddings) of speech sounds that retain linguistically relevant information 
and discard linguistically irrelevant acoustic information like speaker voice type or recording conditions (additive noise, reverberation, etc). In text-based systems, such representations are typically phonemes (as defined by a phonetic dictionary) or characters. Here, we let the representations be latent, which means that they may take any form (dense vectors for each frame, probabilistic codes, discrete codes, etc). This poses an evaluation problem. The ZRC series takes the view that, while discovered units may not necessarily take the shape of straighforwardly linguistically interpretable entities like phonemes or phonetic features, they should at least maintain the same key linguistic \emph{function:} \textbf{linguistic contrast}. Phonemes are typically defined as the smallest element of speech that can induce a potential difference in meaning between words. In English, for instance, the phonemic contrast between /r/ and /l/ is demonstrated by the fact that ``brow'' and ``blow'' are distinct words.  Similarly, here, we ask for representations to distinguish pairs of triphones (``urp,'' ``ulp'') that differ minimally in their center phone, while disregarding variations in speaker or recording condition. We use such triphones that may not necessarily be real words (although they could \textit{potentially} be real words), because the number of actual minimal word pairs in a given corpus is in practice too small to obtain reliable estimates of discriminability between two arbitrary phonemes. Large number of triphone pairs are obtained automatically by mining a speech dataset, and discriminability is computed by running an ABX discrimination test \cite{schatz2016} between the embeddings that a model associates with to these tokens (details in Appendix Section \ref{sec:appendixt1}).



\begin{figure}[]
\centering
\includegraphics[width=1\columnwidth]{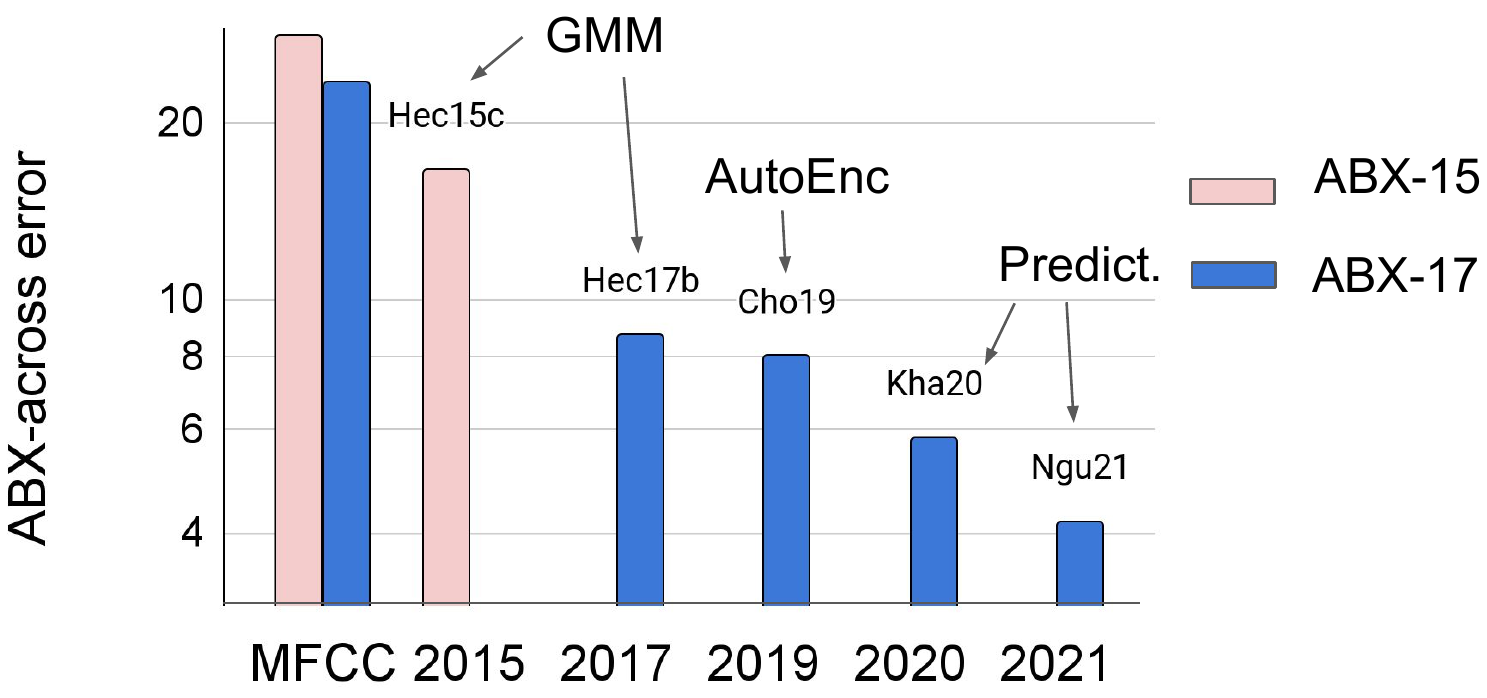}
\caption{ZR Task 1 results on English ABX test sets (ABX-15: Conversational speech-Buckeye; ABX-17: Audiobooks: Librivox). The left two scores are on MFCC representations. The right two scores have been trained on larger datasets (Librispeech 960). See Table \ref{tab:systems} for the references to the systems.}
\label{fig:task1}
\end{figure}

Since 2015, several approaches have been taken. Most start from the basic principle that a central characteristic of text (or phonemic transcription) is that it provides a highly \textit{compressed latent representation} for speech. For reference, a 16~kHz 16-bit waveform is coded using 256 kilobits/sec, which can generic audio codecs like Opus or MP3 can compress to between 32 and 16~Kb/sec (a factor of 8 to 16). In contrast, a phonemic transcription is about 70 \emph{bits}/sec. This represents a compression of more than 200x compared to general audio codecs! The intuition is that phonemes (or text) capture only the linguistically important aspects of speech and discard the rest. Many objective functions proposed for T1 have as their primary goal to reduce the amount of information coded.  A simple and remarkably successful version of this idea---inspired by the ``universal background models'' used in speaker encodings---is to model acoustic frames using a mixture of full-covariance Gaussians. The posteriorgram of the mixture is taken as a new, sparse code for the speech input. This strategy, supplemented with additional speaker normalization tricks, was able to obtain top scores in the ZRC15 and ZRC17 editions (DPGMM \cite{chen15, heck2017feature}: see Figure \ref{fig:task1}). 
Another type of approach seeking to find a compressed latent representation is an autoencoder, which aims at reconstrucing the signal through an information bottleneck, sometimes achieved by using a discrete codebook \cite{badino2015discovering,chorowski2019unsupervised}.
The codebook+wavenet autoencoder of \cite{chorowski2019unsupervised} obtained better results than previous, mixture model-based systems.

Particularly impressive results have been obtained since 2020, when a new generation of models began to appear yielding never-before-seen performance: CPC \cite{oord2018cpc}, wav2vec 2.0 \cite{baevski2020wav2vec}, and HuBERT \cite{hsu2021hubert}; see \cite{lasse} for a review. A salient difference with this new wave of models is that they no longer simply apply the idea of compression, but rather aim to do \textbf{prediction.} In other words, rather than trying to model the probability distribution of acoustic sequences through a compressed latent representation, they aim to predict missing or masked parts of the sequence \textit{conditional} on visible parts of the sequence. For instance, Contrastive Predictive Coding \cite{oord2018cpc} has obtained excellent ABX scores (4.5\% in the across-speaker version) \cite{nguyen2021}. This approach tries to predict future frames (over a window of 10 to 120 milliseconds) based on past ones. Wav2vec 2.0 tries to reconstruct a masked part of the signal (of the order of 100ms), based on left and right context. 
Also novel is these models' \textbf{scale.} They are large, and, accordingly, they are typically trained on large amounts of data (thousands of hours), which is orders of magnitude more than the training sets used in the initial ZRC series. Furthermore, independently of their predictive loss, their \textbf{use of context} is more sophisticated than previous models which integrated context in a finite window \cite{badino2015discovering,thiolliere,chorowski2019unsupervised} or not at all \cite{chen15}. The new models use recurrences, transformers, or both, to constrain unit discovery. Finally, the three models cited above all work \textbf{directly from waveform} instead of relying on engineered features like MFCC or mel filterbanks. (We note that the earlier model of \cite{chorowski2019unsupervised}, which also led to a performance boost, also worked directly from waveform.)

Any or all of these four factors may turn out to be critical to finding good units (or may not). Beyond helping to model temporal correlation at the phonetic level (coarticulation), predictive/masked objectives push models to learn the acoustic properties of units jointly with their functional role in the language---their phonotactics or even lexical or morphological properties---which would be aided by using more sophisticated sequential models. Constraining phonetic discovery by lexical context \cite{naomi11,roland11} or using phonotactics \cite{elsner,ddi} has also showed promise in previous work. 
In addition, allowing models to be large and training them on large amounts of data might push them to mimic the evolution of the human auditory system and its  adaptations to speech. Indeed, \cite{conllj,aclj} find that wav2vec 2.0, HuBERT, and, particularly, CPC, are good predictors of low-level (sub-phonemic) auditory and speech processing in humans. Finally, it is well known that human perception relies on temporal fine structure not captured by magnitude spectrograms  \cite{tfs}, particularly in difficult listening conditions. Models working from waveform might have an advantage. We note, however, that, as \cite{weerts} demonstrate, wav2vec 2.0's use of temporal fine structure still differs radically from that of human listeners.

Other interesting ideas have been explored in the ZRC series. Although they have not made it to the top of the leaderboard, they may still have much to contribute, perhaps in combination with other approaches. 

One idea attempts to capitalize on the idea of using lexical context, but in a very targeted way.  Minimal pairs of real words like ``row'' and ``low'' are rare. While two tokens of the same word will necessarily have the same phonemes, two tokens of different words will on average share only 10\% of their phonemes. The proposals of \cite{thiolliere,zeghidour2016deep} to use Task 2 (Spoken Term Discovery) to constrain acoustic unit discovery by using matched pairs of automatically-discovered words has yielded promising, but not optimal, results---perhaps because unsupervised word discovery is itself difficult (see Section \ref{sec:t2}).

A second idea is to use articulatory or quasi-articulatory information:   given that acoustic units must ultimately allow for the listener learn and pronounce new words (see also Task 4 in Section \ref{sec:t4}), many psycholinguistic approaches propose that human speech processing reconstructs continuous articulatory or discrete quasi-articulatory features  \cite{liberman1967perception,fowler1986event,chomsky1968sound}. The application of this idea to unit discovery by \cite{baljekar2015using} did not yield good results, but, given the importance that sensory--motor coupling has in speech, it seems worthy of further exploration.

Finally, only a few systems have attempted to explicitly model duration. While most models have limited themselves to dealing with short-duration acoustic events (10~ms or so), duration is a principa concern of the HMM-based unit discovery system of \cite{DBLP:conf/sltu/OndelBC16} as well as the segmental CPC approach of \cite{bhati2021unsupervised}. The approach of \cite{murthy2020zero,prakash2020exploration} implicitly considers duration by dividing the signal into syllables, which are then further divided into subsyllabic units (see  also Task 4 in Section \ref{sec:t4}). None of these approaches has reached state-of-the-art performance, but, once again, duration is quite clearly critical to speech perception, and so it seems quite likely that research will need to examine these ideas further.

\subsection{Task 2: Spoken Term Discovery}\label{sec:t2}


Just as the infant learns the words of its language by listening, spoken term discovery seeks to find recurring patterns in the input, proposing a set of boundaries delimiting the start and end of proposed word tokens discovered in the speech, and category labels indicating proposed word types.\footnote{Note here, that contrary to Task 1, we ask systems to explicitly return linguistically interpretable representations (boundaries and cluster labels) instead of a representation that simply satisfies certain functional properties. \textbf{ATTN - est-ce que cette discussion mérite d'être élargie et mise dans la discussion? Ou du moins mis dans le texte principal ici en tant que commentaire sur l'incohérence de la tâche elle-même?}} This problem was explored by several papers prior to the ZRC \cite{park2008,zhang2010,jansen2011,muscariello2012}, and inspired the approach taken to address it as a challenge. However, although the task of ``finding words'' seems intuitively simple, it is made up of at least three subproblems which we evaluate separately.


\begin{itemize}
\item
The \emph{matching} subproblem is to find all pairs of speech fragments that are instances of the same sequence of phonemes. This can be evaluated based on how phonemically similar the discovered fragments are using the gold transcription (normalized edit distance: \textbf{NED}) and how much of the corpus they cover (\textbf{coverage}).

\item The \emph{lexicon discovery} subproblem is to group these fragments into clusters (as opposed to simple pairwise matching). The goal is to find a lexicon of types. A proposed cluster can be evaluated based on how well the members match on the sequence of phonemes (\textbf{Grouping}) and how well the sets match the gold-standard lexicon of word types (\textbf{Type $F$-score}).

\item The \emph{word segmentation} subproblem attempts to find onsets and offsets of fragments that are aligned with the word boundaries as defined in the gold-standard text: we use the \textbf{token} and \textbf{boundary} $F$-scores as is usual in text-based word segmentation.
\end{itemize}

To maximize  comparability with text-based word discovery approaches, all of these evaluations are done at the level of the phonemes obtained forced aligning the test set with its phoneme transcription. Any discovered speech fragment is converted into its transcription, including potentially stranding phonemes on the left or right edge if it contains more than more than 30~ms of that phoneme or more than 50\% of its duration. The full set of scores is presented in detail in Appendix Section \ref{sec:appendixt2}.

\begin{figure}[]
\centering
\includegraphics[width=\columnwidth]{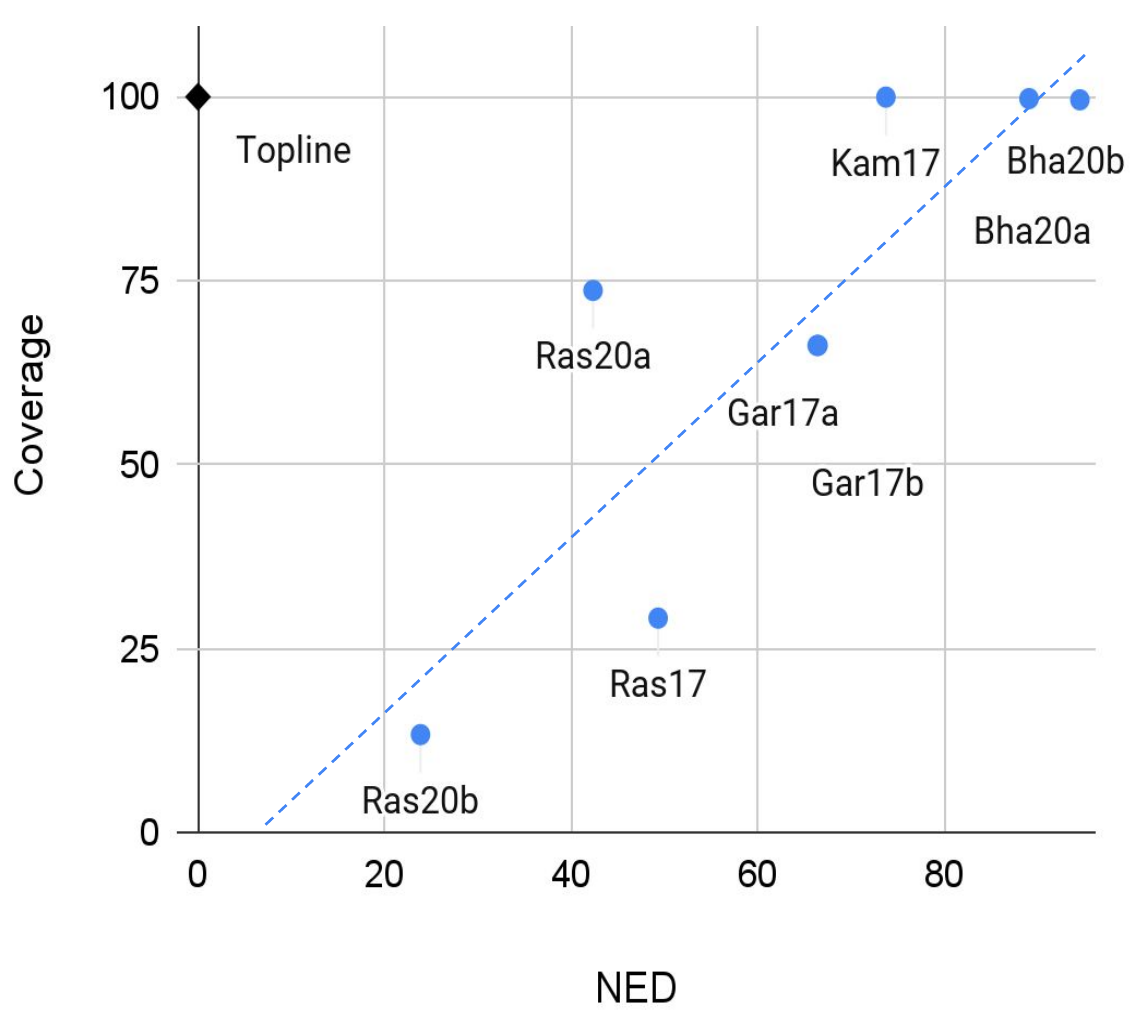}
\caption{Task 3 (term discovery): NED vs Coverage Tradeoff. Average results across 5 languages (ZR17 and ZR20).}
\label{fig:nedcov}
\end{figure}

The Spoken Term Discovery task is still very challenging and has not  received the same attention as Acoustic Unit Discovery. One major finding across the three ZRC editions that featured this task is the existence of a  \textbf{tradeoff} between  attempting to find a lot of words and ensuring that the discovered words are accurate. The quality of the set of words that are treated as matches/repetitions by the system, as measured by the normalized edit distance (NED), will necessarily be better if systems do not commit to extracting more dubious word candidates in the first place; however, the more candidates are ignored, the less of the corpus will receive an analysis (lower coverage) and the fewer of the gold word boundaries will be found (leading also to lower boundary $F$-scores). The tradeoff between term quality and coverage is shown in Figure \ref{fig:nedcov}.


Systems that take a ``matching first'' approach, like \cite{rasanen2015unsupervised,rasanen2020unsupervised}, seek primarily to find recurrent phonetic patterns.  Boundaries are merely designations of the edges of these discovered segments. Such approaches might seem to depend heavily on taking good-quality phonetic representations as input. Indeed, prior to matching,  \cite{rasanen2015unsupervised} does a syllabic segmentation, under the reasoning that such an input representation is more appropriate for word discovery. Nevertheless, \cite{algayres2020evaluating} demonstrated that ABX error rate is not a good indicator of downstream lexicon discovery, and our own informal experiments have shown that naively feeding improved acoustic units into a generic matching system (for example, those learned by \cite{thiolliere}) can actually make matching quality \emph{worse.} The system that currently does best at balancing NED with coverage and segmentation quality takes a  matching-based approach, based only on MFCC inputs \cite{rasanen2020unsupervised}.


Other Task 2 systems take segmentation-oriented approaches, putting a priority on discovering boundaries. None of the systems submitted to the challenge take the extreme approach of searching only for acoustic cues to word boundaries without considering the quality of the spans between boundaries. Rather, building on earlier systems like \cite{lee2015},  systems like  \cite{kamper,kamper2017segmental,kamper2022word} jointly optimize an exhaustive segmentation and a set of  acoustic word embeddings for the segments---segmentations are better if their segments have better word embeddings. Other approaches include the bottom-up approach of \cite{bhati2020self}, using matches to construct a full segmentation, and the joint unit discovery--word segmentation model of  \cite{bhati2021unsupervised}, which posits word boundaries at peaks in surprisal across sequences of learned segmental units. As would be expected, all these systems perform well on segmentation measures. However, among the systems that have been submitted to the full ZR evaluation as of this writing, none come close to \cite{rasanen2020unsupervised} in striking a balance between match quality and coverage.

\begin{figure}[]
\centering
\includegraphics[width=\columnwidth]{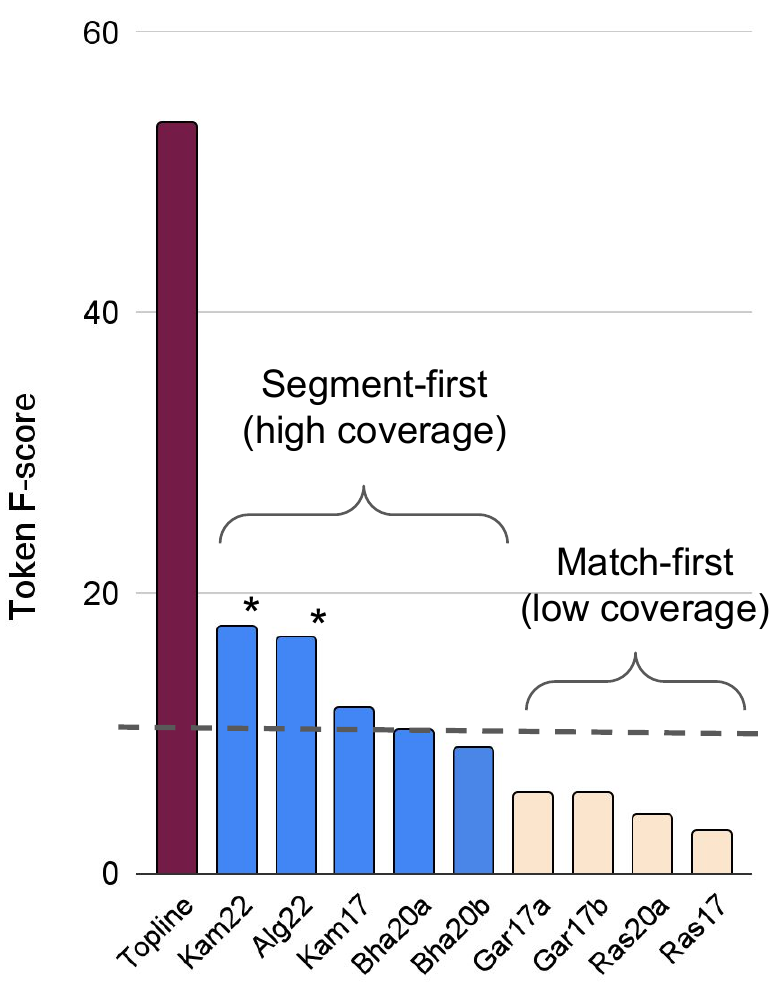}
\caption{Task 3 (term discovery): Token F-scores averaged across 5 languages (ZR17 and ZR 20 plus two new papers). The topline is a unigram word segmentation adaptor grammar trained on the same amount of text. The dotted like is a baseline consisting in random segmentations every 120ms. * these models compute the segmentation without building a lexicon of types.}
\label{fig:tokenFscore}
\end{figure}

Figure \ref{fig:tokenFscore} displays the Token Fscore for each of the submitted systems, compared to a unigram adaptor grammar segmentation system trained on the corresponding text (phonemized text without the blank spaces between words).  All of the high-coverage segmentation-oriented models are on the left and all of the low NED, matching-first models on the  right. The segmentation-oriented models are more likely to do well on this metric, which assesses how many of the word tokens were correctly segmented. As can be seen, the gap between the best speech based models and the text-based ones is still large.

The reason is likely multi-fold. First, speech is variable, which means that the same word will surface with a variety of acoustic shapes. Even if good invariant quantized acoustic representations are used, the potential (and actual) variability in the different ``transcriptions'' in terms of these quantized units for the ``same'' word  grows exponentially with word length. This makes it difficult to build a reliable lexicon of word types. Second, speech rate and phone durations are variable in time, with the result that both phoneme duration and word duration can change subtantially from occurences to occurences, a problem that does not exist in text. Finally, speech is typically coded into frames which has a finer granularity than text (eg, 10ms frames, whereas phonemes last 70ms): a potential word boundary can therefore occur in more places in speech than in text. This increases the number of potential segmentation errors that can be made. This last point is one of the motivations for systems such as \cite{bhati2021unsupervised,cuervo2021contrastive,kamper2022word}, all of which, jointly or sequentially, infer word boundaries hierarchically on the basis of learned acoustic unit boundaries. Any future work will need to address all three of these challenges to achieve better performance in this task. 

\subsection{Task 3: Discrete Resynthesis (TTS without T)}
\label{sec:task3}

\begin{figure*}[]
\centering
\includegraphics[width=2\columnwidth]{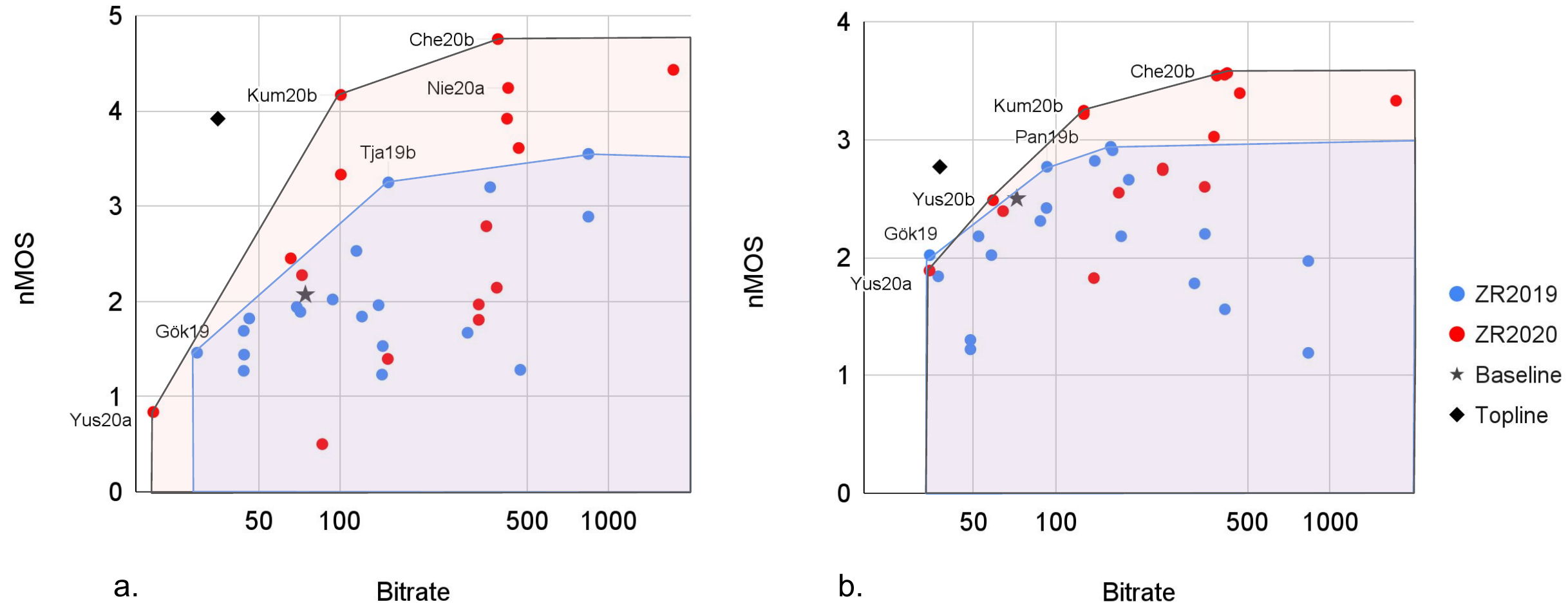}
\caption{ZR Task 3 (TTS WITHOUT T) human evaluation results on the surprise language (Indonesian, panel a) and the dev language (English, panel b) across the two editions of the Challenge (2019 and 2020). The nMOS score is obtained by normalizing the Mean Opinion Score across the two challenges by using the Baseline and Topline as anchor points. }
\label{fig:MOSbitrate}
\end{figure*}

Here, we investigate a task which is similar to what infants may do when they repeat a word or a sentence: they encode the signal into some representation, and then reproduce the same content in their own voice. Defined like this, the task is already known as voice cloning or voice transfer, and it can be performed at a rather low level by introducing a target speaker embedding in the decoder part of a simple encoder-decoder architecture. Here, however, we add the constraint that there be a discrete bottleneck between the encoder and decoder, and we measure the bitrate of the encoding. In other words, we ask  participants to use discovered acoustic units instead of phonemes, and we push these units to approach the bitrate of phonemic transcription.

Prior to the ZRC, \cite{DBLP:conf/icassp/ScharenborgBBHM18}  demonstrated the feasibility of unsupervised discrete resynthesis. Furthermore,  
some of the models in Task 1 \cite{badino2015discovering,chorowski2019unsupervised}  used a similar discrete bottleneck autoencoder architecture, although they did not evaluate the quality of the reconstruction nor the bitrate of the respresentation. 

Participants are provided with a voice training set and a unit training set. The voice data is for building an acoustic model of the target voice for speech synthesis. The unit data contains read speech from multiple speakers (around 10 minutes each). The test dataset consists of novel utterances by unseen speakers, which must be resynthesized in the target voice. Participants submit both the acoustic unit representation and the resynthesis for evaluation.

The general form for the embedding is a sequence of discrete one-hot vectors or vector elements of a finite codebook. In both cases, the bitrate evaluation counts each unique vector value as a single symbol type. The bitrate is calculated as the entropy of the symbols divided by the average duration of the symbols. The quality of the resynthesis is assessed by using human evaluations (Mean Opinion Scores, MOS). For comparison with other tasks, the ABX scores for the discrete codes were also calculated.\footnote{In this challenge, the embedding files do not have any notion of frames or timestamps. This is because we allow participants to submit representations of length shorter than the number of frames, rather than forcing them to duplicate repeated units in the embedding file. Without time alignment, we used a version of ABX in which the systems are fed the short waveforms corresponding to each test triphone, rather than, as we do elsewhere, feeding them entire utterances from which the embeddings for the triphones is extracted using the time alignments. This may  penalize systems that have not been trained to encode very short speech segments.}       




The performance on the downstream task was overall quite good, with several systems achieving better resynthesis than the text-based top-line. As shown in Figure \ref{fig:MOSbitrate}, there is a general tradeoff between synthesis quality and bitrate, which held both in the Dev language (English) and in the heldout test language (Indonesian). As shown by the black point in the figure (the decoded output of a simple phone recognizer), phonemic transcription is a highly-compressed representation of speech which is excellent for this task (the middling MOS scores are attributable to the fact that we did not optimize our out-of-the-box ASR and TTS systems to the task).

Many of the systems that have a very low bitrate (consistently under 100 bits/sec)  learn a discrete autoencoder on acoustic features \cite{eloff2019unsupervised,yusuf2019temporally,liu2019unsupervised,gundogdu2020vector}, generally then taking  further steps such as filtering or downsampling to reduce the temporal resolution.  On the other hand, our baseline model, as well as the related work of \cite{yusuf2021hierarchical}, learn latent HMMs as acoustic units, in order to explicitly model duration. On the other hand, \cite{nayak2019virtual,prakash2020exploration}  puts its temporal reduction in an initial step of acoustic segmentation based on syllable-like units. Among these models, \cite{prakash2020exploration}, which presegments and then learns HMM acoustic units, stands out as reaching performance comparable to higher bitrate models (though it admittedly has a higher bitrate than the other models listed). Syllable-like presegmentation, as noted above, is also the current most effective strategy for optimizing the tradeoff between match quality and coverage in Task 2. It is fair to say that this approach has been underutilized, given its promise.

Most of the remaining systems have a bitrate between 100 and 600. Supervised posteriorgrams are on the upper end of this, and MFCC representations have a bitrate around 1500.\footnote{Note that our bitrates are calculated empirically on a speech corpus. The representational \emph{capacity} of MFCC representations is an order of magnitude higher.} Most of the submitted systems in this range are compression approaches using discrete autoencoders, including the system of \cite{chen2020unsupervised}, which gives excellent performance. The system of \cite{van2020vector} stands out among the others as yielding high quality results. This is the only submitted system which uses a predictive loss based on CPC---although, unlike typical CPC models, it works from spectrogram and is trained on the  small (15h) dataset provided for the 2019 edition.

The results of \cite{lakhotia2021generative} also support the claim that CPC and related approaches are well-adapted to discrete resynthesis (in addition to supporting entirely new tasks through their capacity to be used as language models: see Section \ref{sec:gslm} below). Independently,  \cite{lakhotia2021generative} demonstrated that an automatic evaluation using ASR is strongly predictive of human evaluators' ratings.

\subsection{Task 4: spoken LM}

Spoken language modeling is the task of learning a language model directly from the audio. In Task 4, we do not presuppose the input to the language model. The inputs could be discrete or continuous representations from Task 1 or word level representations from Task 2, so long as these representations are learned without supervision from text or other labels. The task can be understood as the modeling of the probability distribution of spoken utterances in an unknown language.

Here, the evaluation challenge is severe. Language models trained from text are typically evaluated by the perplexity of a given corpus as computed by the language model, or by fine-tuning on downstream tasks. However, as discussed above, we 
restrict our evaluation to zero-shot tasks that require no training. This excludes a fine tuning evaluation. As for perplexity, in text-based systems, it is derived from the conditional probabilility distribution of the next token given a past sequence of tokens. In speech-based systems that use discrete pseudo-text units, the number of such units is a latent variable, making the perplexities difficult to compare across models. The problem becomes worse for systems that do not use discrete representations at all, where the estimation of the conditional probabilities themselves become model dependent. 
Instead of computing an average perplexity across a corpus, we propose to use a contrastive approach, where we compare a `surprisal score' (negative log probability, or simply the loss function itself) for minimal pairs of utterances, one legal, the other illegal. We verify whether the value of the surprisal score is higher for the legal than for the illegal utterance. We use this logic to probe the lexical level (words versus nonwords) and the syntaxic level (grammatical versus non grammatical sentences), using stimuli constructed using  the Wuggy nonword generator \cite{keuleers2010wuggy} and derived from the BLIMP dataset \cite{warstadt2019blimp} (see \cite{nguyen2020zero} and Appendix Section \ref{sec:appendixt4}). 

Additionally, language models are also sometimes evaluated by probing tasks, which investigate the nature of the representations computed in the hidden units. We adapt a semantic similarity test  previously used for evaluating text-based word embeddings to correlate the similarity of systems' representations of words with human similarity judgments. This enables us to measure the extent to which the model is able to extract lexical semantic knowledge. 

Inspired by the idea that young children learn in a multimodal, multisensory enviroment, some earlier models of word discovery and representation learning proposed to use videos or images in training \cite{harwath2016unsupervised,arandjelovic2017look,chrupala2017representations}. In this vein, \cite{alishahi2021zr} proposed to extend all of Task 4 to such ``visually-grounded'' training. Here, the task challenge is quite difficult. While (most) language learners have access to a rich visual environment, the correlation between speech and audio is relatively weak: spoken image captions, for example, are very far from being a realistic input. Yet, not only do realistic audio-visual datasets that accurately simulate the environment of the infant not exist, it is not even clear exactly what properties these datasets should have \cite{chrupala2022visually,nikolaus2022learning}. Rather than let this stop us from exploring the question, however, we have now given Task 4 participants the option of indicating that they used visual grounding in training, but, in part for this reason, Task 4 differed from previous instances of the challenge in removing the requirement to train on any specific data set---participants indicate the dataset they used. Systems are all submitted to the same, speech-only evaluation at test time.

Task 4 was opened for submissions for the first time in early 2021, and the initial round of submissions was documented in \cite{zr2021summary}. The visually grounded option was only introduced afterwards. Table \ref{tab:res2021b} presents the results of the latest submissions to Task 4 (those submitted since \cite{zr2021summary}, including the visually grounded systems. 

\newcommand{\gr}{\rowcolor[HTML]{E0E0E0}}
\begin{table*}[ht]
\caption{\textbf{Leaderboard update for December 2021.} The first three systems are \textbf{reference systems,} including the two best systems submitted before the publication of \cite{zr2021summary}. The first three newly submitted systems are \textbf{visually grounded}. The remaining submitted systems use strictly self-supervised audio training. The best overall score in each column among all submitted systems is bolded.\label{tab:res2021b}}
\centering
\begin{tabular}{cccccccccc} 
&    &  \multicolumn{2}{c}{ABX-with.} &  \multicolumn{2}{c}{ABX-across}& sWUGGY & sBLIMP  & \multicolumn{2}{c}{sSIMI}   \\
\cline{3-4}\cline{5-6}\cline{9-10}
System  & Budget & clean  & other & clean  & other  & & & synth.   & Libri. \\ \hline\hline
Character-based (RoBERTa) & 24576       & -      & -     & -      & -     & {96.25}   & {82.11}  & {23.22}  & {21.08}  \\
\gr Best March Low-budget  \cite{bn} & 60 &5.41  &{8.67}&{6.89}&{13.14}&{72.86} &{53.59}  &{5.60}&2.72 \\
    Best March high-budget \cite{bert-baseline} & 60 &3.28 &4.81&{4.31}&{7.92}&{75.51} &{56.16}  &3.19&1.32\\ \hline\hline
\gr Pen21  (VG) \cite{pengnew} & 468  & 4.24 & 5.22 & 5.08 & 7.91 & 75.23 & 57.40 & \textbf{17.99} & \textbf{12.78}\\
    Lee21a (VG) & 7  & 6.50 & 9.95 & 9.17 & 15.46 & 51.91 & 50.43 & 4.51 & 5.54\\
\gr Lee21b (VG) & 7  & 6.50 & 9.95 & 9.04 & 15.44 & 51.91 & 50.43 & 8.1 & 12.21\\ \hline
    Gan21  & 60  & 5.04 & 7.89 & 7.08 & 14.01 & 59.87 & 52.20 & 7.47 & 4.38\\
\gr Ngu21a & 2560  & 3.49 & 5.72 & 4.56 & 9.19 & 77.22 & 55.62 & 8.43 & 4.48\\
    Ngu21b & 2112  & 3.80 & 4.86 & 4.70 & 7.70 & 79.13 & 58.89 & 8.62 & 6.65\\
\gr Ngu21c & 3424  & {3.03} & \textbf{3.62} & 3.83 & \textbf{5.63} & 80.19 & 59.29 & 6.8 & 3.08\\
    Ngu21d & 2720  & 3.49 & 5.72 & 4.55 & 9.20 & \textbf{80.29} & \textbf{59.93} & 9.43 & 5.33\\
\gr Lan21  & 60  & \textbf{2.85} & 4.44 & \textbf{3.67} & 7.33 & 70.23 & 51.93 & 3.09 & 2.05\\
    Bha21a & 60  & 3.49 & 5.72 & 4.51 & 9.16 & 50.90 & 52.10 & 3.44 & 1.67\\
\gr Bha21b & 60  & 3.49 & 5.73 & 4.57 & 9.19 & 48.55 & 52.80 & 1.57 & 5.2\\
    Gao21a & 60  & 6.71 & 10.62 & 8.41 & 15.16 & 62.29 & 53.17 & 1.67 & 1.13\\
\gr Gao21b & 60  & 6.60 & 10.86 & 8.51 & 16.51 & 68.42 & 54.02 & 2.61 & 1.78\\
    Gao21c & 35  & 7.45 & 11.76 & 10.94 & 19.31 & 62.29 & 53.17 & 5.07 & 1.2\\
\hline
\end{tabular}
\end{table*}

Task 4 is clearly in its very early stages (this in spite of the excellent ABX performance of the units used in systems up to now). However, even at this stage (after only one year's worth of submissions) spoken language modelling has shown improvement on the lexical decision task (moving from the best speech-based baseline's 75\% accuracy up to 80\%) and on the syntactic judgment task (improving from 56\%  to 60\% accuracy). The approach so far has been simple: high-quality units and a powerful language model. In our baseline models as well as most submissions, these components were trained separately; newer models like HuBERT \cite{hsu2021hubert} learn them jointly. The two approaches are currently tied for the top position on the leaderboard. As for capturing word semantics, the Fast-VGS+ system of \cite{harwath22} stands out as a serious competitor. This visually-grounded system takes advantage of spoken image caption data in training.


\section{What next?}

\subsection{Problems and paradoxes}

Text-like representations---be they phonemic, alphabetic, or logographic writing---are fundamental to speech and language processing. On the one hand, they record the linguistically important properties of the form (what was actually uttered). On the other hand, text-like representations support straightforward analysis of the content (``meaningful'' properties like morphology, syntax, and semantics).

The discreteness of text plays an important role in relating the two. While listeners are sometimes sensitive to detailed, sub-phonemic properties \cite{gwilliams2018spoken}, and while various kinds of gradations in lexical meaning can be observed \cite{beekhuizen2021probing}, the two kinds of variability are not generally correlated: although it is possible to pronounce the noun \emph{sun} with an initial sound that sounds a bit more like the adjective \emph{fun,} this change does not evoke the concept of a star which is any more amusing or make the word any more susceptible to be used as an adjective. Text-like representations would thus seem to be \emph{necessary} for the pipeline laid out in Figure \ref{fig_1}; and, apart from prosody and paralinguistic information, where text is inadquate, text-like representations appear to be for \emph{sufficient} syntactic and semantic tasks.


 Given this assumption, and given that the ABX phone discrimination measure was originally intended to serve as a rough indicator of how similar the set of discovered units was to the sound inventory of the language---and thus, in a loose sense, how text-like---it is with some consternation that we have gradually discovered that the link between the ABX evaluation and performance on other tasks is complicated---an object of study in itself.
 
 Task 3 is a case where our assumption proves both useful and problematic. The ABX score is strongly correlated with the intelligibility of speech synthesis starting from those units \cite{dunbar2019zero,dunbar2020}, and a simple supervised ASR baseline fits nicely onto the line of best fit. 
 However, as discussed in Section \ref{sec:task3}, good performance on Task 3 does not require text-like units. The phonemic transcription remains an outlier among very low bitrate units in terms of the quality of synthesis it produces. Otherwise, higher bitrate units---thus \emph{less} discrete and presumably less text-like---yield better synthesis.   There is clearly something special and desirable about the phonemic transcription, but this does not mean that it is the optimal representation for speech resynthesis. More importantly, higher bitrate units tend to  have better ABX scores.
 
 This is \emph{not} to say that representations that capture all acoustic detail are necessarily ``good units'' by the standards of the ABX score. The phone ABX score demands units that are invariant to phone and to speaker---the MFCC baseline was beat in the very first ZR challenge. However, the space of possible low bitrate units is large. Lower bitrate units must make decisions about what information to throw out---decisions that can be risky if not done exactly right. In fact, this can even be seen in our own  results in \cite{dunbar2019zero}: contrary to our typical topline supervised posteriorgrams, which show good ABX scores, our topline for Task 3 was a phonemic \emph{decoding} from an  ASR system (we did this for comparability with the discrete units submitted to Task 3). The system was taken out-of-the-box and not optimized for the task. Consequently,  the decodings yielded very poor ABX scores---in English, they were worse than the MFCCs! Given that the task of simply decoding low-bitrate units is hard, it is  unsurprising that \emph{discovering} good low-bitrate units remains difficult.
 
Furthermore, as discussed above, the subtask of matching discovered word candidates in Task 2 has not, up to now, shown clear benefits from using pretrained units, and the related
unsupervised token frequency estimation task of \cite{algayres2020evaluating} is only weakly correlated with the phone ABX score of the underlying units (0.53). 
Similarly, the sWuggy (spot-the-word) measure of Task 4 can also be taken as an assessment of how useful the underlying units are for building a lexicon, and the correlation between this measure and the ABX score is similarly middling (0.52).

Thus, in sum, the intuition that ``better'' and more ``text-like'' units will render tasks such as synthesis and word discovery easier is not clear-cut. This leads us to ask whether the ABX score itself needs improvement as a measure of unit quality. Several properties not directly measured currently are phoneme-level (rather than phone-level) invariance, invariance under time distortion, and robustness to noise.

On the other hand,  a useful alternative criterion for the quality of subword units might directly measure their ability to discriminate between matched versus unmatched pairs of words, rather than phones. This poses an evaluation challenge, however, as the unmatched pairs selected for evaluation must be selected carefully so as to be actually difficult for the systems in question, rather than trivially easy (see also \cite{riad2018sampling}).

\subsection{Continuous challenge and evaluations 2.0}

As of press time, the submission site is open continuously and has been reworked to make it easy to run the evaluations on all of the past benchmarks and add the newest results. ZeroSpeech events (special sessions, workshops) will be continue to be announced regularly at conferences and in the community, particularly as new tasks and evaluations are added.

Some of the new evaluations planned for the near future are improvements to the unit quality measures along the lines discussed above (including integration of human speech perception benchmarks such as \cite{millet2020perceptimatic,millet2021predicting}), and automation of the speech synthesis quality measures that are currently done by human evaluators.
The spoken language modelling task is in its infancy, and there are a number of levels of linguistic analysis that are not included (for example, morphology). This is the case more generally for the challenge (for example, we have never attempted to evaluate prosody). 

%
%
%
%





\subsection{A new task: Generative spoken language modelling}
\label{sec:gslm}

Combining Task 3 and Task 4 leads us to consider the possibility of \emph{generating} spoken language. A traditional spoken dialogue system will (conditional on some knowledge source) generate text, typically using a neural language model. The text is then synthesized into speech. A spoken language model, can be made to generate spoken language directly, as demonstrated by \cite{lakhotia2021generative,kharitonov2021data}. Much as Task 3 is complementary to Task 1---but has slightly different constraints---the task of generating speech from a spoken language model will be developed into a Task 5.

\subsection{Towards interactive and cross-modal learning}

The 2021 challenge experimented with allowing visually grounded models---an experiment that appears to have paid off. This kind of weak supervision is capable of greatly improving the scores on the language modelling tasks. This raises more questions than it answers, as the precise nature of the visual stimuli (video, labelled images, non-parallel data), may matter a lot to the result. Future iterations may develop more targeted sub-challenges aimed at understanding what kind of multimodal data may be useful for learning speech representations, and why.

In addition to access to other sensory modalities, another aspect of typical human language development which has conspicuously absent from the challenge is the learner's interaction with the environment---both with other people, and in manipulating the material environment. While this aspect is challenging to model, it may nevertheless prove to be important for any autonomous system learning from speech. A broad framework for how to do modelling of language learning in the context of social interaction is given in \cite{tsuji2021}.

\section{Conclusion: Towards textless NLP}
 
Research on unit discovery has led to a wave of new models using unsupervised pre-training to advance ASR in both high-resource and low-resource languages. Yet, the possibility of using less text, or non-parallel text, while promising, has done little to change the dominance of text as the basic currency of NLP pipelines. One notable exception is translation, in which the idea of completely dispensing with text and training machine translation directly from speech to speech in end-to-end fashion has seen substantial uptake. In general, however, converting speech to text and back remains the first and the last step of  current speech-based NLP systems.  

The allure of replacing text with low-bitrate representations of speech discovered in an unsupervised way goes beyond bringing NLP to lower-resource languages. Learned pseudo-text promises to be more flexible than traditional orthography: if the transcription system is the result of learning, then it can also change to deal with new varieties and accents; it can also learn from the start to capture important linguistic information which is captured in a very limited way by typical writing systems, such as prosody. On the other hand, it also promises to be more consistent: many writing systems are complicated by arbitrary exceptions, while other languages lack standardized conventions for spelling. Discovered representations should avoid both of these issues. Unlike traditional phonetic transcription, which uses a fixed, universal set of symbols which can actually have rather different phonetic values across languages, unit discovery allows for a system to be adapted to the language (though, as seen above, some of the better approaches lean heavily in the direction of building a universal representational system which can then be fine-tuned to specific languages).

The Zero Resource Speech Challenge has spearheaded efforts to build demonstrably useful unit discovery, as well as stimulating  progress (albeit modest) in applying these representations to more complex tasks. The major advances in building more realistic auditory-like representations have already borne fruit in recognition and synthesis. As we move toward an emphasis on unit robustness, we look ahead to the possibility of truly textless NLP---and a major key to unlocking cognitive models of human language development and speech perception.

\section*{Acknowledgment}

For E. Dunbar, this research was supported by  the Connaught Fund and  the Arts and Science Tri-Council Bridging Fund, University of Toronto, and by French Agence Nationale de la Recherche grants ANR-17-CE28-0009 (GEOMPHON),  ANR-11-IDEX-0005 (USPC),  and ANR-10-LABX-0083 (EFL). For E. Dupoux in his EHESS role \& N. Hamilakis, this work was supported by French Agence Nationale de la Recherche grants ANR-10-IDEX-0001-02 PSL* and ANR-19-P3IA-0001 PRAIRIE 3IA Institute, and by a Meta AI Research Gift.


\section{Appendix}\label{sec:appendix}

Here, we provide details about the metrics used in the ZRC series.

\subsection{Task 1 Metrics: within and across speaker ABX}\label{sec:appendixt1}
The minimal pair ABX task \cite{schatz2013,schatz2014} does not require any training and only requires to define a dissimilarity metric between speech tokens.  It is inspired by match-to-sample tasks used in human psychophysics and measures discriminability between two sound categories. 
We define the ABX-discriminability of category $A$ from category $B$ as the probability that tokens $a$ and $x$ from $A$ are further apart than token $b$ from $B$ and $x$ are, according to a dissimilarity function $d$. The discriminability score is symmetrized by averaging of the ABX discriminability of $A$ from $B$ and of $B$ from $A$. If the tokens are representated at a frame level, dynamic time warping is used to realign the two tokens, and frame-level dissimilarities are averaged along the realignment path to obtain $d$. Most submissions used angular dissimilarity (arccos of the normalized dot product of the frame embeddings), while others used the KL divergence when the frame representations are posterior probabilities. 

In all of iterations of the SRC series so far, categories $A$ and $B$  are sequences of 3 phonemes that differ in the central sound (not necessarily real words, e.g., ``beg''--``bag'', ``api''--``ati'', etc). The compound measure sums over all minimal pairs of this type found in the corpus in a structured manner, that depends on the task. For the
\textit{within-speaker} task, all of the phone triplets belong to the same speaker (e.g. $A=\textrm{beg}_{T1}$, $B=\textrm{bag}_{T1}$, $X=\textrm{bag}'_{T1}$). The scores for a given minimal pair are first averaged across all of the speakers for which this minimal pair exists. The resulting scores are then averaged over all found contexts for a given pair of central phones (e.g. for the pair /a/-/e/, average the scores for the existing contexts such as /b\_g/, /r\_d/, /f\_s/, etc.). Finally the scores for every pair of central phones are averaged and subtracted from 1 to yield the reported within-talker ABX error rate. For the \textit{across-speaker} task, $A$ and $B$ belong to the same speaker, and $X$ to a different one. $A=\textrm{beg}_{T1}$, $B=\textrm{bag}_{T1}$, $X=\textrm{bag}_{T2}$. The scores for a given minimal pair are first averaged across all of the pairs of speakers for which this contrast can be made. As above, the resulting scores are averaged over all contexts over all pairs of central phones and converted to an error rate.

\begin{table}[]
\caption{\textbf{Characteristics of the different ZRC ABX benchmarks.} \label{tab:abx_bench}}
\begin{tabular}{llll}
\hline
Benchmark & Language (dataset)  & Type & Duration/Speakers   \\
\hline
ABX-15    & \begin{tabular}[c]{@{}l@{}}English (Buckeye)\\ Xitsonga ()\end{tabular}                            & \begin{tabular}[c]{@{}l@{}}conversations\\ Timit-like\end{tabular}                                 & \begin{tabular}[c]{@{}l@{}}5h, 12 spk\\ 2h30, 24 spk\end{tabular} \\ \cline{2-4}
ABX-17    & \begin{tabular}[c]{@{}l@{}}English (Librivox)\\ French (Librivox)\\ Mandarin (xxx) \\ German (L1) (Librivox)\\ Wolof (L2) (??)\end{tabular} 
          & \begin{tabular}[c]{@{}l@{}}audiobook\\ audiobook\\ read speech\\ audiobook\\ Timit-like\end{tabular} & \begin{tabular}[c]{@{}l@{}}27h\\ 17h\\ 25h\\ 11h\\ 5.9h\end{tabular} \\  \cline{2-4}
ABX-LS    & English (Librispeech)  & audiobook   & \begin{tabular}[c]{@{}l@{}}dev/test clean/other:\\ 5h each\end{tabular} \\
\hline
\end{tabular}
\end{table}

\newcommand{\set}[1]{\left\{#1\right\}}
\newcommand{\tup}[1]{\langle#1\rangle}
\newcommand{\cdisc}{C_{\textrm{disc}}}
\newcommand{\pdisc}{P_{\textrm{disc}}}
\newcommand{\fdisc}{F_{\textrm{disc}}}
\newcommand{\fall}{F_{\textrm{all}}}
\newcommand{\pall}{P_{\textrm{all}}}
\newcommand{\pgold}{P_{\textrm{all}}}
\newcommand{\psubs}{P_{\textrm{disc*}}}
\newcommand{\pclus}{P_{\textrm{clus}}}
\newcommand{\bdisc}{B_{\textrm{disc}}}
\newcommand{\bgold}{B_{\textrm{gold}}}
\newcommand{\pgoldclus}{P_{\textrm{goldclus}}}
\newcommand{\pgoldlex}{P_{\textrm{goldLex}}}
\newcommand{\fgoldlex}{F_{\textrm{goldLex}}}
\newcommand\SetB[3]{\ensuremath{\{\text{\ensuremath{#1 \mid} \parbox[t] {\widthof{\ensuremath{#3}}} {\ensuremath{#2}\}}}}}
\newcommand\SetA[2]{\ensuremath{\{\text{\ensuremath{#1 \mid} \parbox[t] {\widthof{\ensuremath{#2}}} {\ensuremath{#2}\}}}}}
\newcommand{\cover}[1]{\mathrm{cover}(#1)}
\newcommand{\flatten}[1]{\mathrm{flat}(#1)}
\newcommand{\nmatch}[1]{\mathrm{occ}(#1)} 
\newcommand{\weight}[1]{\mathrm{weight}(#1)}
\newcommand{\types}[1]{\mathrm{types}(#1)}
\newcommand{\freq}[1]{\mathrm{freq}(#1)}
\newcommand{\ned}[1]{\mathrm{ned}(#1)}
\newcommand{\card}[1]{\left\vert{#1}\right\vert}

\subsection{Task 2 Metrics}\label{sec:appendixt2}
The evaluation of spoken term discovery systems as matching systems consists of two scores,  \textbf{NED} and  \textbf{coverage}. \textbf{NED,} is the average, over all matched pairs, of the Levenshtein distance between their phonemic transcriptions, divided by the max of their phonemic length.

The second score is the \textbf{coverage,} the fraction of the discoverable part of the corpus that is covered by all the discovered fragments. The discoverable part of the corpus is found by computing the union of all of the intervals corresponding to all of the pairs of ngrams (with n between 3 and 20). This is almost all of the corpus, except for unigram and bigram hapaxes.

Six scores are used to evaluate the performance of a spoken term discovery system in terms of lexicon discovery. The first three are \textbf{grouping precision, recall} and \textbf{F-score.} These are defined in terms of $\pclus$, the set of all pairs of fragments that are groups in the same cluster, and $\pgoldclus$, the set of all non-overlapping pairs of fragments which are both discovered by the system (not necessarily in the same cluster) and have exactly the same gold transcription. 

{\small
\begin{align}
 \mbox{Prec:} & \sum_{t\in\types{{\pclus}}} w(t,\pclus)\frac{\card{\nmatch{t, \pclus\cap\pgoldclus}}} {\card{\nmatch{t,\pclus}}} \label{eq:grouping:precision} \\
 \mbox{Rec:} & \sum_{t\in\types{{\pgoldclus}}} w(t,\pgoldclus)\frac{\card{\nmatch{t, \pclus\cap\pgoldclus}}} {\card{\nmatch{t, \pgoldclus}}}\label{eq:grouping:recall}
\end{align}
}

Where $t$ ranges over the types of fragments (defined by the transcription) in a cluster $C$, and $occ(t,C)$ is the number of occurrences of that type, $w$ the number of occurrences divided by the size of the cluster. In other words, Prec is a weighted measure of cluster purity and Rec, of the inverse of the cluster's fragmentation. The other three scores are \textbf{type precision, recall,} and \textbf{F-score}. Type precision is the probability that discovered types belong to the gold set of types (real words), whereas type recall is the probability that gold types are discovered. We restrict both sets to words between three and twenty segments long.

\subsection{Task 3 Metrics}\label{sec:appendixt3}

\textit{Intelligibility} was measured by asking participants to orthographically transcribe the synthesized sentence. Each transcription was compared with the gold transcription using the Levenshtein distance, yielding a Character Error Rate (\textit{CER}).  The overall \textit{naturalness} of the synthesis was assessed on a 1 to 5 scale, yielding a Mean Opinion Score (\textit{MOS}).\footnote{The question posed was: \emph{Rate how natural the audio is, between  1 and 5 (1=very unnatural, 3 = neutral, 5=very natural).}} \textit{Speaker similarity} was assessed using a 1 to 5 scale. Sentences were presented in pairs (target voice, system voice).\footnote{The question posed was: \emph{Rate the similarity between the reference
 voice and the system voice, between 1 
and 5 (1 = very different voices,
 3 = neither similar nor different voices,
 5 = very similar voices).} Ten additional trials were included, for each participant, in which the reference voice was not the target voice but the source voice.} A  training phase occurred before each task. Three ``catch'' trials were included in the transcription, consisting of easy sentences from the original corpus not included in the rest of the experimental list, allowing us to detect participants that  failed to do the task.

Each participant performed the evaluation tasks in the same order (Intelligibility, Naturalness, Similarity), the overall evaluation lasting about one hour.  To avoid re-evaluation of the same sentence by the same participant, the sentences (types) were split into two disjoint subsets: one third for the Intelligibility task (62 for English, 49 for Indonesian), and two third for the Naturalness task (129 for English, 100 for Indonesian). The complete set of sentences was used in the Similarity task. In the Intelligibility and Naturalness tasks, all the sentences were seen by all subjects; in the Similarity task, a pseudo random one-third of the whole sentences was selected for each participant. Each sentence token was evaluated at least once with each system (the submitted, topline and baseline systems, as well as the original recordings).\footnote{In the Intelligibility task, each system was evaluated at least 70 times for English, and 148 times for Indonesian, with each combination of sentence and system seen at least once. In the Naturalness task, each system was evaluated at least 180 times for English, and 274 for Indonesian, with each combination of sentence and system seen at least 36 times for English and at least 68 times for Indonesian. In the Similarity task, each system was evaluated at least 89 times in English, and 120 times in Indonesian, and all possible combinations of sentence and system were seen by at least one participant.} English judges were recruited through Mechanical Turk. Indonesian judges were recruited through universities and research institutes in Indonesia. All were paid the equivalent of 10 USD. Only data from participants with $<$0.80 CER on catch trials were retained (Development: 35/35; Surprise: 68/69).

For the \textit{bitrate} computation, each vector is processed as a character string. A dictionary of the possible values is constructed over the embedding file for the submitted test set. 
We thus assume that the entire test set corresponds to a sequence of vectors $U$ of length $n$: $U=[s_1,...,s_n]$. The bit rate for $U$ is then $B(U)=n \sum_{i=1}^{n}{\frac{p(s_i)log_{2}p(s_i)}{D}}$, where $p(s_i)$ is the probability of symbol $s_i$. The numerator is $n$ times the entropy of the symbols, which gives the optimal number of bits needed to transmit the sequence of symbols $s_{1:n}$. To obtain a bitrate, we divide by $D$, the total duration of $U$ in seconds.\footnote{A fixed frame rate transcription may have a higher bitrate than a ``textual'' representation due to the repetition of symbols across frames. For instance, the bitrate of a 5 ms framewise gold phonetic transcription is around 450 bits/sec and that of a ``textual'' transcription around 60 bits/sec.}

 \subsection{Task 4 metric}\label{sec:appendixt4}
\textbf{Acoustic: the Libri-light ABX metrics.} The ABX metric consists in estimating, for two speech categories $A$ and $B$ (e.g., `bit' and `bet'), the probability that two exemplars $x$ and $a$ of the same category $A$ are closer to one another than two exemplars $x$ and $b$ of different categories $A$ and $B$. 
The score is aggregated across all pairs of triphones like `bit' and `bet', where the change occurs in the middle phoneme. This can be computed within-speaker ($a$, $b$ and $x$ are from the same speaker) or across-speaker ($a$ and $b$ are from the same speaker, and $x$ from a different speaker). To compute this score, participants are required to provide an embedding for each input triphone and to specify a pseudo-distance between acoustic tokens. By default, we provide such a distance, which is the average along a Dynamic Time Warping path realigning $a$, $b$ and $x$ of a distance between embedding frames (angular distance). This metric is agnostic to the dimensionality of the embeddings, can work with discrete or continuous codes, and has been used to compare ASR speech features \cite{schatz2016abx}.  Here, we run it on the pre-existing Libri-light dev and test sets, which has been already used to evaluate several self-supervised models \cite{Kahn_2020_librilight,riviere2020unsupervised}.

\textbf{Lexicon: the sWUGGY spot-the-word metrics.} 
In this task, the models are presented with a pair of spoken tokens: an existing word and a matching nonword.  Participants are to provide a number (probability or pseudo-probability) associated to each acoustic tokens, and models are evaluated on their average accuracy of word-nonword classification based on this probability (chance level: 0.5). The sWUGGY test and development sets consists of 20,000 and 5,000 pairs respectively, with the existing words being part of the LibriSpeech train vocabulary. We also prepared additional OOV-sWUGGY test and development sets consisting of 20,000 and 5,000 pairs respectively, with existing words which do not appear in the LibriSpeech training set. The nonwords are produced with WUGGY \cite{keuleers2010wuggy}, which generates, for a given word, a list of candidate nonwords best matched in phonotactics and syllabic structure, which we additionally filtered for pronouncability using G2P, and for having on average the same unigram and bigram phoneme frequencies as words. Stimuli were produced with the Google Speech API. 

\textbf{Syntax: the sBLIMP acceptability metrics.}  
This part of the benchmark is adapted from BLIMP \cite{warstadt2019blimp}, a set of linguistic minimal sentence pairs of matched grammatical and ungrammatical sentences. Similarly to sWUGGY, the task is to decide which of the  two is grammatical based on the probability of the sentence. The test and dev sets contain 63,000 and 6,300 sentence pairs respectively, with no sentence pair overlap. Stimuli were filtered to contain LibriSpeech vocabulary and for natural prosodic contours, and synthesised as above.

\textbf{Lexical Semantics: the sSIMI similarity metrics.} Here, as in \cite{chung2018speech2vec}, the task is to compute the similarity of the representation of pairs of words and compare it to human similarity judgements. As for the ABX task, participants provide embeddings for input tokens as well as a distance to compute similarity. Here, we provide by default the cosine distance computed over pooled embeddings (with mean, max or min pooling). We used a set of 13 existing semantic similarity and relatedness tests: WordSim-353 \cite{yang2006verb},
WordSim-353-SIM \cite{agirre2009study}, mc-30 \cite{miller1991contextual}, rg-65 \cite{rubenstein1965contextual}, Rare-Word (or rw) \cite{luong2013better}, simLex999 \cite{hill2015simlex},
simverb-3500 \cite{gerz2016simverb}, verb-143 \cite{baker2014unsupervised} , YP-130 \cite{yang2006verb}
and the relatedness-based datasets include MEN \cite{bruni2012distributional}, Wordsim-353-REL \cite{agirre2009study}, mturk-287 \cite{radinsky2011word}, and mturk-771 \cite{halawi2012large}.
All scores were normalised on a 0-10 scale, and pairs within a same dataset containing the same words in different order were averaged.  Pairs containing a word absent from LibriSpeech train set \cite{Panayotov2015librispeech} were discarded. We  selected as development set the mturk-771 dataset and the other 12 datasets were used as test sets, making sure that no pair from the development set was present in any of the test sets.

We then created two subsets of audio files, one synthetic (using the Google API), one natural obtained by retrieving the audio extracts from  LibriSpeech corresponding to each word, following the process presented in \cite{chung2018speech2vec}. In this subset, each word can appear in multiple tokens, providing phonetic diversity; duplicated scores are averaged in the analysis step. The natural subset is smaller than its synthesised counterpart, as we had to discard pairs from the test and dev sets which were not present in the LibriSpeech test and dev sets respectively.  The synthesized subset is composed of 9744 and 705 word pairs for the test and dev sets respectively, and the LibriSpeech subset is composed of 3753 and 309 pairs for the test and dev sets.

\subsection{List of contributed systems}

In Table \ref{tab:systems}, we list the ID of the submitted system, and their corresponding bibliographic references.

\begin{table}[]
\caption{List of contributed systems}\label{tab:systems}
\begin{tabular}{lrlll}
\hline
System  & Date & Paper                     & Github & Description \\
\hline
\multicolumn{5}{c}{\underline{\it ABX-15 (Task 1- Eng. \& Xit)}}                      \\
Bad15a--c& 2015 &   \cite{badino2015discovering} &        &             \\
Bal15a--e& 2015 &  \cite{baljekar2015using} &        &             \\
Che15a--d& 2015 &  \cite{chen2015parallel}                        &        &             \\
Fah15   & 2015 &  \cite{myrman2017partitioning}   &        &             \\
Hec15a--d& 2015 &   \cite{heck2016unsupervised}  &        &             \\
Ren15a--d& 2015 &   \cite{renshaw2015comparison} &        &             \\
Sri15a,b& 2015 &  \cite{srivastava2016articulatory}  &        &             \\
Thi15   & 2015 &  \cite{thiolliere2015hybrid}   &        &             \\
Zeg15   & 2015 &  \cite{zeghidour2016deep} &        &             \\
\hline
\multicolumn{5}{c}{\underline{\it ABX-17 (Task 1- Eng. \& Xit)}}                      \\
Ans17a--d& 2017 &  \cite{ansari2017deep}                        &        &             \\
Che17a,b& 2017 &   \cite{chen2017multilingual}   &        &             \\
Hec17a,b& 2017 &  \cite{heck2017feature}  &        &             \\
Pel17a,b& 2017 &   \cite{pellegrini2017technical}                       &        &             \\
Ras17   & 2017 &  Unpub       &        &             \\
Shi17a,b& 2017 &    \cite{shibata2017composite}   &        &             \\
Yua17a-c& 2017 &  \cite{yuan2017extracting}    &        &             \\
Cho19   & 2019 &   \cite{chorowski2019unsupervised}    &        &             \\
Kha20   & 2020 &  \cite{kharitonov2021data}      &        &             \\
\hline
\multicolumn{5}{c}{\underline{\it TDE-15 (Task 2- Eng. \& Xit.)}}                      \\
Baseline& 2015 &  \cite{jansen2011efficient,versteegh_2015}       &        &             \\
Kam15a  & 2015 & \cite{kamper2017embedded}    &        &             \\
Lyz15a-c& 2015 & \cite{lyzinski2015evaluation}   &        &             \\
Ras15a-c& 2015 &                          &        &             \\
\hline
\multicolumn{5}{c}{\underline{\it TDE-17 (Task 2- 5 languages)}}                      \\
Baseline& 2017 &                          &        &             \\
Kam22  & 2022 &\cite{kamper2022word}     &        &             \\
Alg22   & 2022 &\cite{alg22}              &        &             \\
Kam17  & 2017 &\cite{kamper2017segmental}&        &             \\
Bha20a,b& 2020 &                          &        &             \\
Gar17a,b& 2017 &                          &        &             \\
Ras20a,b& 2020 &                          &        &             \\
Ras17   & 2017 &                          &        &             \\
\hline
\multicolumn{5}{c}{\underline{\it TTS0-19 (Task 3- English, Indonesian)}} \\
Baseline& 2019 &                          &        &              \\
Tja19a,b& 2019 &                          &        &              \\
Hor19a,b& 2019 &                          &        &              \\
Pan19a,b& 2019 &                          &        &              \\
Kam19a,b& 2019 &                          &        &              \\
Fen19a,b& 2019 &                          &        &              \\
Ral19   & 2019 &                          &        &              \\
Cho19a,b& 2019 &                          &        &              \\
Yus19   & 2019 &                          &        &              \\
Liu19a,b& 2019 &                          &        &              \\
Kum19a,b& 2019 &                          &        &              \\
Gok19   & 2019 &                          &        &              \\
Yus20a-c& 2020 &                          &        &              \\
Hou20a,b& 2020 &                          &        &              \\
Mor20   & 2020 &                          &        &              \\
Kum20a,b& 2020 &                          &        &              \\
Che20a,b& 2020 &                          &        &              \\
Lum20a,b& 2020 &                          &        &              \\
Nie20a,b& 2020 &                          &        &              \\
Tja20a,b& 2020 &                          &        &              \\
\hline
\multicolumn{5}{c}{\underline{\it sLM-20 (Task 4- English)}} \\
Baseline& 2021 &                          &        &              \\
Pen21   & 2021 &                          &        &              \\
Lee21a,b& 2021 &                          &        &              \\
Gan21   & 2021 &                          &        &              \\
Ngu21a-d& 2021 &                          &        &              \\
Lan21   & 2021 &                          &        &              \\
Bha21a  & 2021 & \cite{scpc}              &        &              \\
Gaoa-c  & 2021 &                          &        &              \\
\hline
\end{tabular}

\end{table}

\bibliographystyle{IEEEtran}
\bibliography{zr}

\begin{IEEEbiography}[{\includegraphics
[width=1in,height=1.25in,clip,
keepaspectratio]{dupoux.png}}]
{IEEE Publications Technology Team}
In this paragraph you can place
your educational, professional background
and research and other interests.
\end{IEEEbiography}

\begin{IEEEbiography}[{\includegraphics
[width=1in,height=1.25in,clip,
keepaspectratio]{dupoux.png}}]
{IEEE Publications Technology Team}
In this paragraph you can place
your educational, professional background
and research and other interests.
\end{IEEEbiography}

\begin{IEEEbiography}[{\includegraphics
[width=1in,height=1.25in,clip,
keepaspectratio]{dupoux.png}}]
{Emmanuel Dupoux}
is a Professor at the Ecole des Hautes Etudes en Sciences Sociales (EHESS) and Research Scientist at Meta AI Labs. He directs the Cognitive Machine Learning team at the Ecole Normale Supérieure (ENS) in Paris and INRIA.  His education includes a PhD in Cognitive Science (EHESS), a MA in Computer Science (Orsay University) and a BA in Applied Mathematics (Pierre & Marie Curie University). His research mixes developmental science, cognitive neuroscience, and machine learning, with a focus on the reverse engineering of infant language and cognitive development using unsupervised or weakly supervised learning. He is the recipient of an Advanced ERC grant, co-organizer of the Zero Ressource Speech Challenge series (2015--2021) and the Intuitive Physics Benchmark (2019) and led in 2017 a Jelinek Summer Workshop at CMU on multimodal speech learning. He is a CIFAR LMB and a ELLIS Fellow. He has authored 150 articles in peer reviewed outlets in cognitive science and language technology. 
\end{IEEEbiography}